\pdfoutput=1

\documentclass[11pt]{article}

\usepackage[preprint]{acl}

\usepackage{times}
\usepackage{latexsym}

\usepackage[T1]{fontenc}

\usepackage[utf8]{inputenc}

\usepackage{microtype}

\usepackage{inconsolata}

\usepackage{graphicx}
\usepackage{subcaption}
\usepackage{caption}

\usepackage{balance}

\usepackage{amsmath}
\usepackage{amssymb}
\usepackage{mathtools}
\usepackage{amsthm}
\usepackage{algorithm}
\usepackage{algorithmic}
\usepackage{xcolor} 
\usepackage{colortbl} 

\usepackage{enumitem}
\usepackage{svg}
\usepackage{booktabs}
\usepackage{multirow}
\usepackage{makecell}
\newcolumntype{C}[1]{>{\centering\arraybackslash}m{#1}}

\usepackage[most]{tcolorbox}
\usepackage{dashrule}
\usepackage{wrapfig}

\usepackage{pifont}
\usepackage{minted}
\usepackage{changepage}

\usepackage[capitalize,noabbrev]{cleveref}
\crefname{section}{Sec.}{Secs.}
\Crefname{section}{Section}{Sections}
\Crefname{table}{Table}{Tables}
\crefname{table}{Tab.}{Tabs.}

\newcommand{\ours}[0]{PilotRL\ }

\title{PilotRL: Training Language Model Agents via Global Planning-Guided Progressive Reinforcement Learning}

\author{Keer Lu$^{\ddagger}$, Chong Chen$^\mathsection$, 
Bin Cui$^{\ddagger}$, Yunhuai Liu$^{\ddagger}$, Wentao Zhang$^{\dagger}$ \\
$^\dagger$Center for Data Science, Academy for Advanced Interdisciplinary Studies, Peking University, \\ 
$^\mathsection$Huawei Cloud BU, 
$^\ddagger$School of Computer Science, Peking University \\
keer.lu@stu.pku.edu.cn, 
chenchong55@huawei.com, \\ 
\{bin.cui, yunhuai.liu, wentao.zhang\}@pku.edu.cn \\
}

\definecolor{codebg}{rgb}{0.95, 0.95, 0.95} 
\definecolor{codeborder}{rgb}{0.7, 0.7, 0.7} 

\begin{document}

\addtocontents{toc}{\protect\setcounter{tocdepth}{0}}

\maketitle

\begin{abstract}
Large Language Models (LLMs) have shown remarkable advancements in tackling agent-oriented tasks. 
Despite their potential, existing work faces challenges when deploying LLMs in agent-based environments. 
The widely adopted agent paradigm ReAct centers on integrating single-step reasoning with immediate execution, which limits its effectiveness in complex tasks requiring long-term strategic planning. 
Furthermore, 
the coordination between the planner and executor during problem-solving is also a critical factor to consider in agent design. 
Additionally, 
current approaches predominantly rely on supervised fine-tuning, 
which often leads models to memorize established task completion trajectories, thereby restricting their generalization ability when confronted with novel problem contexts. 
To address these challenges, 
we introduce an adaptive global plan-based agent paradigm \textbf{AdaPlan}, aiming to synergize high-level explicit guidance with execution to support effective long-horizon decision-making. 
Based on the proposed paradigm, 
we further put forward \textbf{PilotRL}, a global planning-guided training framework for LLM agents driven by progressive reinforcement learning. 
We first develop the model’s ability to follow explicit guidance from global plans when addressing agent tasks. 
Subsequently, based on this foundation, we focus on optimizing the quality of generated plans.
Finally, we conduct joint optimization of the model's planning and execution coordination. 
Extensive experiments indicate that \textbf{PilotRL} could achieve state-of-the-art performances, 
with LLaMA3.1-8B-Instruct + PilotRL surpassing closed-sourced GPT-4o by 3.60\%, while showing a more substantial gain of 55.78\% compared to GPT-4o-mini at a comparable parameter scale. 
\end{abstract}

\section{Introduction}
\label{sec:intro}

An \textit{agent} can be defined as an entity capable of perceiving its environment, making decisions, and executing actions in pursuit of predefined or adaptive goals~\citep{wooldridge1995intelligent,maes1995agents,jennings1998roadmap}. 
The state-of-the-art Large Language Models (LLMs), such as GPT-4~\citep{achiam2023gpt} 
and Gemini~\citep{team2023gemini}, 
have exhibited strong agent capabilities, including instruction following, reasoning, and programming, which inspires widespread efforts to develop autonomous agent systems with LLMs serving as central cognitive controllers~\citep{song2023llm,sumers2023cognitive}. 
Nevertheless, considering the high financial costs and safety risks of close-sourced proprietary models~\citep{li2023multi,yuan2023gpt}, recent efforts have been shifted to improve such agent capabilities in open-sourced models as effective alternatives~\citep{chen2024agent,song2024agentbank,fuagentrefine}.

\begin{figure}[t]
    \centering
    \vspace{-1em}
    \includegraphics[width=1\linewidth]{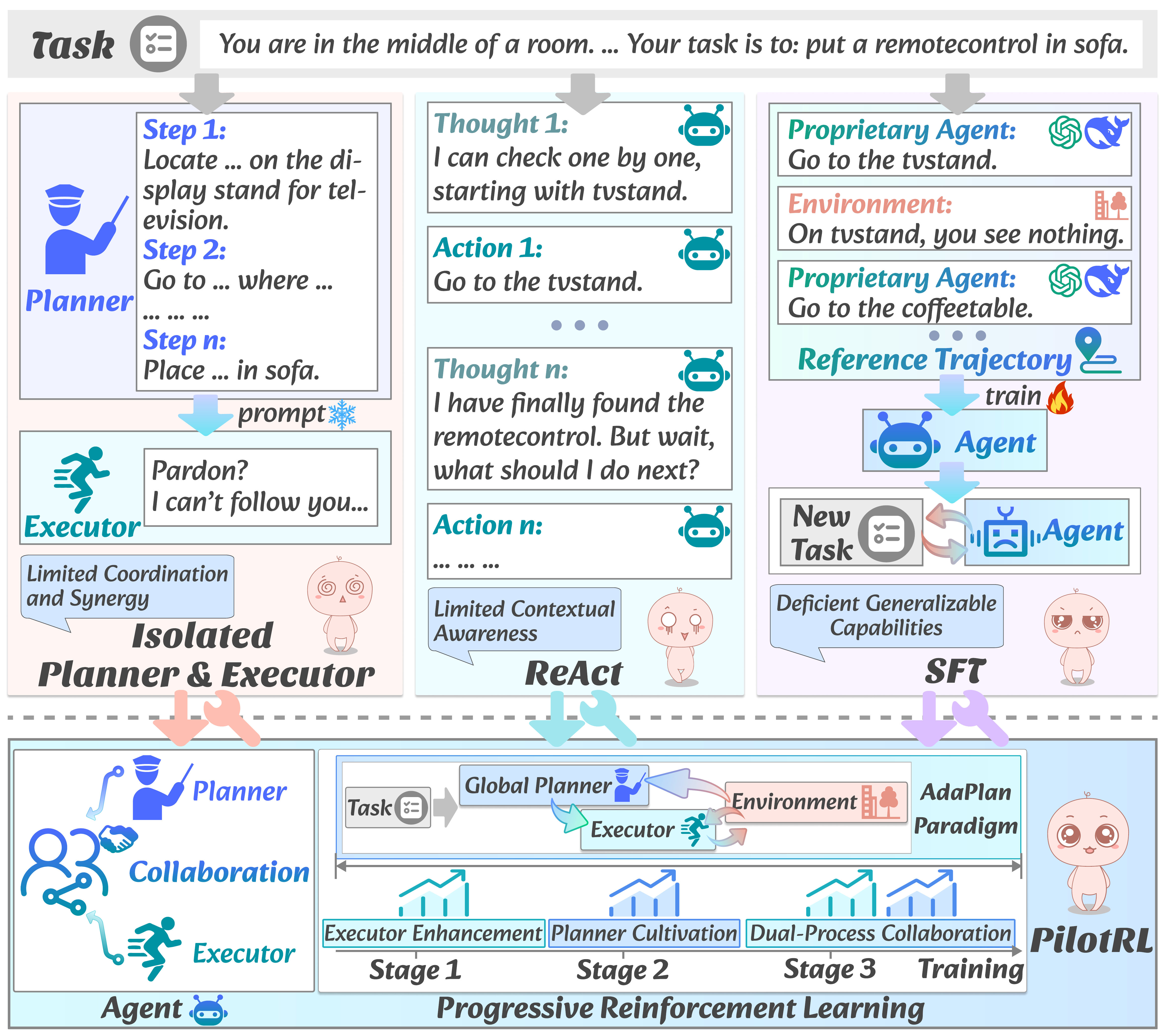}
    \caption{Comparison of \ours (\textit{bottom}) with existing methods (\textit{top}) for agent task completion.}
    \label{fig:intro}
    \vspace{-1em}
\end{figure}

Despite their potential, existing works face some limitations, 
as shown in \Cref{fig:intro}: 
\textit{\textbf{(C1) Limited Contextual Awareness of ReAct}}: 
While the ReAct paradigm~\citep{yao2023react} is a general foundation of agentic systems, it lacks insight into the overarching context. 
The reasoning component (generated as ``thought'') focuses purely on immediate action, which limits its effectiveness in complex tasks requiring sequential execution. 
\textit{\textbf{(C2) Insufficient Coordination between Planning and Executing}}: 
Although recent studies have incorporated planning into agent-based 
problem-solving~\citep{erdogan2025plan,xiong2025mpo}, they design the planner and executor in isolation, leading to potential mismatches between the two components. As a result, 
the generated plans 
may not be effectively followed by the executor, undermining overall task performance. 
\textit{\textbf{(C3) Deficient Generalization of SFT}}: 
Extensive research has been devoted to enhancing the agent capabilities of models through supervised fine-tuning (SFT)~\citep{deng2023mind2web,zeng2024agenttuning}. 
However, studies indicate that SFT tends to lead models to memorize task-specific heuristics rather than acquiring generalizable capabilities applicable to new scenarios~\citep{chu2025sft}.

To address these challenges, we introduce \textbf{\textit{PilotRL}}, 
a global plan-driven reinforcement learning framework for the training of LLM agents. 
For \textit{\textbf{C1}} and \textit{\textbf{C2}}, 
we propose the adaptive global plan-based paradigm \textit{AdaPlan} to guide the agent through complex tasks as a pilot, 
where global plans are dynamically generated and continuously updated throughout the execution process. 
The global planner and executor are implemented within a unified model to enhance their coordination and mutual adaptability.
For \textit{\textbf{C3}}, 
we employ reinforcement learning (RL) for its high effectiveness at enhancing generalizable knowledge in LLMs~\citep{jaech2024openai,guo2025deepseek,team2025kimi}, 
the training process of which can be divided into three stages: 
\textbf{(1) Stage 1: Executor Enhancement.} 
We begin by developing the executor’s instruction adhesion to the global plan when addressing agent tasks. 
\textbf{(2) Stage 2: Global Planner Cultivation.} 
Building upon the global plan following capabilities acquired in Stage 1, we subsequently optimize the global planner to improve the quality of generated plans. 
\textbf{(3) Stage 3: Joint Optimization.} 
Finally, we refine the coordination between the global planning and execution of models to enhance their collaborative performance in agent scenarios.

\textbf{Contributions.} 
The main contributions can be summarized as follows: 

\begin{itemize}[leftmargin=*]
    \item \textit{\underline{Paradigm Innovation.}} 
    We introduce an adaptive global plan-based agent paradigm, AdaPlan,
    to synergize high-level reasoning with executing for long-horizon decision-making. 
    By integrating both the global planner and executor in a unified model, our approach enables more effective coordination and improved end-to-end performance. 
    \item \textit{\underline{Training Framework Advancement.}} 
    Based on AdaPlan, 
    we propose PilotRL, a global planning-guided progressive reinforcement learning framework designed for enhancing the agent capabilities of models via a three-stage process. 
    \item \textit{\underline{Performance and Effectiveness.}} 
    Experiments indicate the superiority of PilotRL. 
    Notably, models trained with \ours even surpasses closed-sourced proprietary models for agent tasks, achieving average improvements over GPT-4o and GPT-4o-mini by 2.35\% and 53.90\%. 
\end{itemize}


\section{PilotRL}
\label{sec:method}

\begin{figure*}[t]
    \centering
    \includegraphics[width=1\textwidth]{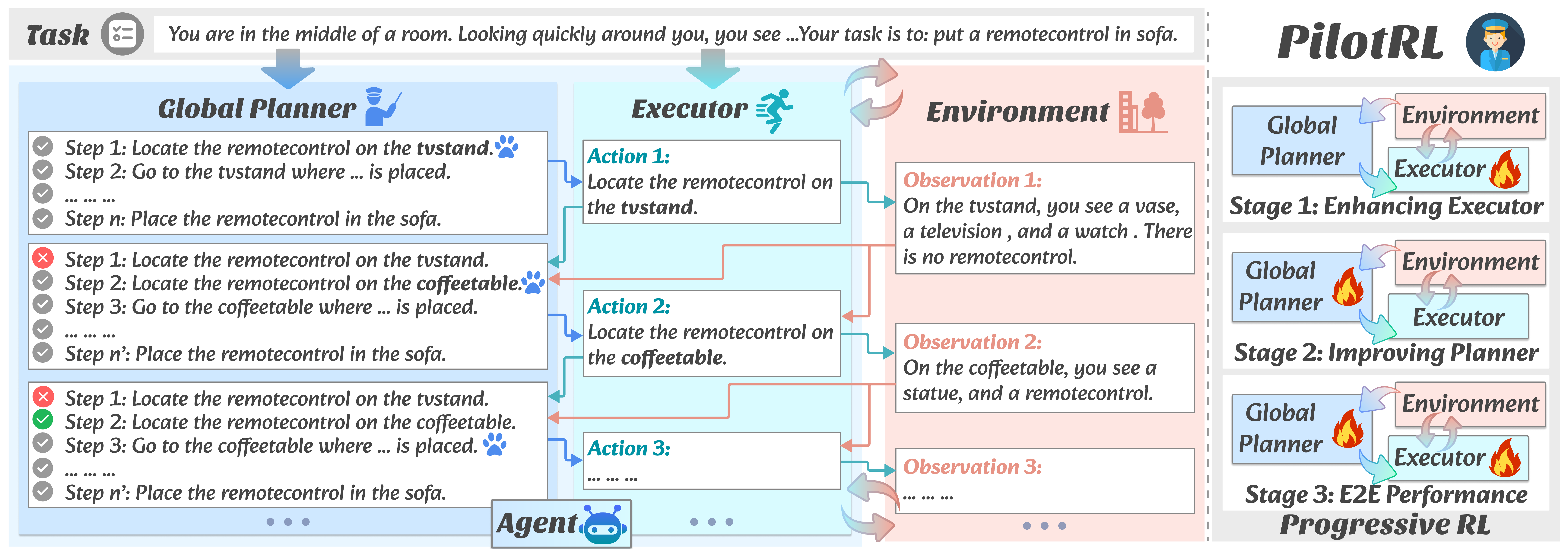}
    \caption{Overview of \textbf{\textit{PilotRL}}. 
    (\textit{Left}) In \textbf{\textit{AdaPlan}} paradigm, the \textit{global planner} begins by processing the task instruction and generates an initial high-level plan for guidance, which is then passed to the \textit{executor} for action generation. 
    The observation from the \textit{environment} is then fed back to both the \textit{executor} for subsequent action generation and the \textit{global planner} for plan adaptation in response to changes or unexpected outcomes. 
    (\textit{Right}) The three-stage training process of our \textbf{\textit{progressive Reinforcement Learning (RL)}}. 
    }
    \label{fig:pipeline}
    \vspace{-1em}
\end{figure*}

Assuming the scenario where an agent interacts with an environment for task solving, we present a detailed overview of 
our \ours framework.

\subsection{AdaPlan: Adaptive Global Planning}

While ReAct~\citep{yao2023react} is effective in many interactive agent tasks, its 
reliance on 
single-step reasoning and immediate action generation limits its capability in scenarios that require extended planning and coherent decision-making.
To address this, 
we introduce the \textbf{\textit{AdaPlan}} paradigm, 
which focuses on the adaptively generated and refined global plan throughout the task-solving process.

As shown in the left part of \Cref{fig:pipeline}, the agent architecture consists of two key components: the \textit{global planner} and the \textit{executor}. 
For a given task instruction $G$ and the initial context $\mathcal{C}^{(0)}$, the global planner first generates the global plan $\mathcal{P}^{(0)} = [p_1^{(0)}, p_2^{(0)}, ..., p_{N_0}^{(0)}]$ consisting of $N_0$ steps, where $p_i^{(0)}$ represents the recommended action for the executor at step $i$ under the current planning strategy. 
At each time step $t$ ($t \ge 1$), the executor takes an action $a^{(t)} \in \mathcal{A}$ based on: \textbf{\textit{(1)}} the previous accumulated context 
$\mathcal{C}^{(t-1)}=\{(a^{(j)}, o^{(j)})\}_{j=0}^{t-1}$, where $o \in \mathcal{O}$ refers to observation from the environment, and 
\textbf{\textit{(2)}} the guidance from the current global plan $\mathcal{P}^{(t-1)}$.  
Subsequently, it receives the resulting observation $o^{(t)} \in \mathcal{O}$, and the current turn of agent-environment interaction $(a^{(t)}, o^{(t)})$ is incorporated into the accumulated context $\mathcal{C}^{(t)}$ of the execution step $t$. 
The global plan $\mathcal{P}^{(t-1)}$ is then iteratively refined according to the task goal $G$ and the accumulated context $\mathcal{C}^{(t)}$ to facilitate the next execution, resulting in $\mathcal{P}^{(t)}$. Each $p_i^{(t-1)}$ in the original $\mathcal{P}^{(t-1)}$ is updated as follows: 
\begin{gather}  
    p_i^{(t)} = 
    \begin{cases}  
        p_i^{(t-1)}, \text{if}\, i \le t \\  
        \pi (p_i^{(t)} | G, \mathcal{C}^{(t)}, \mathcal{P}^{(t-1)}, i), \text{if}\, i > t &
    \end{cases}
    \label{eq:global_plan_adaptation}    
\end{gather}
where $\pi$ is the adaptation policy of the global plan generator. 

By dynamically updating the global plan based on real-time feedback from executor-environment interactions, the agent can promptly assess the validity and efficiency of the current planning strategy and make necessary adjustments accordingly. Furthermore, in cases where the executor deviates from the prescribed plan, the 
global planner 
can adaptively revise the course of action to guide the executor toward more effective task execution.

\subsection{Progressive Reinforcement Learning}

Within the global planning-driven agent paradigms, two key factors influence the overall performance: the quality of the generated global plans, and the degree to which the executor adheres to the plan’s directives when interacting with the environment. 
Accordingly, we employ a three-stage pipeline for training, as shown in the right part of \Cref{fig:pipeline}. 

\subsubsection{Stage 1: Enhancing the Instruction Adherence of Executor} 
\label{subsubsec:training_stage1}

The ability to comply with the guidance of the global planner is foundational to the entire agent paradigm. 
Therefore, our focus lies on improving the executor's capacity to follow existing global plans as well as acquire a thorough understanding of the action space $\mathcal{A}$ in the initial training stage. 
Here we utilize the frontier model, e.g., DeepSeek-V3~\citep{liu2024deepseek}, for the provision of each global plan. 
Specifically, for each time step $t$: 
\begin{itemize} 
    \item \textbf{Plan Generation}: We first prompt the model to generate all possible global plans based on the specified task goal $G$ and accumulated contextual information $\mathcal{C}^{(t)}$ to provide a comprehensive set of potential candidates. 
    \item  \textbf{Plan Selection}: Following this, the model evaluates each of these candidate plans across the dimensions of correctness, executability, format validity, etc., and selects the most suitable one to guide the executor’s actions. 
\end{itemize}

In general, the reward metric in Stage 1 is \textit{the sum of normalized} components: \textbf{\textit{format}}, \textbf{\textit{adherence degree}}, and \textbf{\textit{end-to-end (E2E) performance}}.

\textit{\textbf{Format.}} \quad 
The model is required to produce its outputs according to a predefined output paradigm. Specifically: 
\textit{\textbf{(1)}} All responses should be chosen from the two actions, ``Thought'' or ``Action'', and must strictly align with the formats of ``Thought: ... Action: ...'' or ``Action: ...''. 
\textit{\textbf{(2)}} The output must be produced in a readable format, without distorted or illegible characters, and then the environmental feedbacks are encapsulated within \texttt{<observation>...</observation>} tags. 
Based on the above requirements, the format reward is: 
\begin{gather}  
    \mathcal{R}_{{format}} = 
    \begin{cases}  
        1, \text{if the format is correct} \\  
        0, \text{if the format is incorrect} &
    \end{cases}
    \label{eq:stage1_format}
\end{gather}

\textbf{\textit{Adherence Degree.}} \quad 
This aspect constitutes a core component in fostering the executor’s compliance with the global plan during Stage 1. 
Here we employ a frontier model (e.g., DeepSeek-V3) as the evaluator to score the generated actions. It assesses whether the model’s output semantically aligns with the current step of the global plan. Actions are assigned a score of 2 for fully compliant, 1 for partially compliant (e.g., for suggested actions that require the invocation of multiple tools, at least one tool is utilized to support the execution), and 0 for noncompliant actions: 
\begin{gather}
\resizebox{0.89\linewidth}{!}{
    $\mathcal{R}_{adherence} = 
    \begin{cases}  
        2, \text{if completely compliant} \\  
        1, \text{if partially compliant} \\
        0, \text{if noncompliant} &
    \end{cases}$
}
\label{eq:stage1_adherence}
\end{gather}

\textbf{\textit{End-to-End (E2E) Performance.}} \quad 
The measurement of the first two components concentrates solely on individual execution, rather than assessing the holistic interaction between the agent and the environment. 
However, in real-world interactions, the problem-solving process may exhibit trajectory redundancy or unintended topic drift, leading to unpredictable deviations from the intended workflow. Therefore, it is essential to obtain a comprehensive, end-to-end view of agent performance in order to assess whether the current interaction trajectory aligns with the expected behavior, and to ensure that the target task is accomplished efficiently and directly, without unnecessary detours. 
\begin{gather}
\resizebox{0.89\linewidth}{!}{
    $\mathcal{R}_{E2E} = 
    \begin{cases}  
        2, \text{if accomplished efficiently} \\  
        1, \text{if accomplished with redundancy} \\
        0, \text{if unaccomplished} &
    \end{cases}
    \label{eq:stage1_e2e}$
}
\end{gather}
We use DeepSeek-V3 to evaluate the end-to-end performance $\mathcal{R}_{E2E}$. 
The agent-environment interactions receive a score of 2 if the task is accomplished in a direct and efficient manner without process redundancy. A score of 1 is given if the task is completed but the interaction involves trajectory redundancy or topic drift. If the agent fails to achieve the objective, it is assigned a score of 0.

\subsubsection{Stage 2: Cultivating the Capacity of Global Planner}

Following the initial training stage, the agent has acquired a foundational paradigm for global plan following and action execution. 
In this stage, we shift our focus to enhancing the agent’s ability to generate global plans.  
In generating the global plan, we adopt a \textit{generate-then-select} strategy similar to that used in Stage 1 with the frontier model, 
which enhances the quality of the global plan ultimately used for explicit guidance, leading to more effective and coherent decision-making. 
Specifically, all feasible global plans that could potentially solve the given task are first generated, and then the most appropriate one is selected from this pool of candidates. 
The reward function design in Stage 2 
is \textit{the sum of normalized} components: 
\textbf{\textit{format}}, \textbf{\textit{end-to-end (E2E) performance}}, and \textbf{\textit{global plan quality}}, with the first two already formally defined in \Cref{eq:stage1_format} and \Cref{eq:stage1_e2e}. 

\textbf{\textit{Global Plan Quality}} \quad 
When evaluating the quality of the generated global plan, we consider three primary dimensions: \textit{correctness}, \textit{executability}, and \textit{standardization}. 
\textbf{\textit{(1)}} \textit{Correctness} assesses whether the plan effectively leads to the fulfillment of the task objectives. 
\textbf{\textit{(2)}} \textit{Executability} evaluates the clarity and ease with which the agent can adhere to the instructions, as indicated by the alignment of the executor's action with the global planner’s directives. 
\textbf{\textit{(3)}} \textit{Standardization} checks whether the generated instructions conform to a consistent and well-defined format. 
The quality score of the global plan is calculated as follows: 
\begin{equation}
    \mathcal{R}_{planning} 
    = R_{correct} + R_{execute} + R_{standard}
    \label{eq:stage2_planning}    
\end{equation} 
where components $R_{correct},R_{execute},R_{standard} \in \left \{ x \in \mathbb{Z} \mid 1 \leq x \leq 5 \right \}$, 
with 5 indicating the best performance. 
We use the frontier model DeepSeek-V3 as the evaluator to score each dimension. 

\subsubsection{Stage 3: Orchestrating the End-to-End (E2E) Performance}

Having separately enhanced the model’s capabilities in 
both generating and complying with global plans 
in earlier stages, 
Stage 3 focuses on strengthening the coordination between 
the global planner and the executor, 
i.e., the joint optimization of our global planning-driven agent paradigm \textit{AdaPlan}. 
The reward function at this stage is \textit{the sum of normalized} \textbf{\textit{format}} and \textbf{\textit{end-to-end (E2E) performance}}, which directly prioritizes comprehensive performance of the ultimate task objective.

\section{Experiments}
\label{sec:experiments}

\subsection{Experimental Setup}
\label{subsec:experimental_setup}

\begin{table*}[tb]
    \vspace{-4mm}
    \centering
    \resizebox{\textwidth}{!}{
        \begin{tabular}{c|c|cccccccc}
            \toprule
                \multirow{2}{*}{Backbone Model} & \multirow{2}{*}{Method} & \multirow{2}{*}{w/o Plan.} & \multicolumn{1}{|c}{ALFWorld} & IQA & TextCraft & Wordle & \multicolumn{1}{|c}{BabyAI} & \multicolumn{1}{c|}{MAZE} & \multirow{2}{*}{\textbf{Avg.}} \\
            \cline{4-9}
             & & & \multicolumn{4}{|c}{\textit{In-Domain (ID)}} &  \multicolumn{2}{|c|}{\textit{Out-of-Domain (OOD)}} & \\
            \midrule
            \multicolumn{10}{c}{\cellcolor[HTML]{EFEFEF}\textbf{\textit{Close-Sourced Models}}} \\
            \midrule
            \multirow{1}{*}{GPT-4o} & -- & \ding{55} & \multicolumn{1}{|c}{75.83} & 66.59 & 68.50 & 78.65 & \multicolumn{1}{|c}{57.87} & 60.42 & \multicolumn{1}{|c}{67.98} \\
            \multirow{1}{*}{GPT-4o-mini} & -- & \ding{55} & \multicolumn{1}{|c}{52.35} & 40.32 & 46.74 & 42.51 & \multicolumn{1}{|c}{43.96} & 34.36 & \multicolumn{1}{|c}{45.21} \\
            \midrule
            \multicolumn{10}{c}{\cellcolor[HTML]{EFEFEF}\textbf{\textit{Open-Sourced Agent-Specific Models}}} \\
            \midrule
             \multirow{1}{*}{Agent-FLAN-7B} & -- & \ding{55} & \multicolumn{1}{|c}{70.54} & 57.62 & 24.66 & 22.28 & \multicolumn{1}{|c}{24.39} & 28.93 & \multicolumn{1}{|c}{38.07} \\
            \multirow{1}{*}{LLaMA-xLAM-2-8B-fc-r} & -- & \ding{55} & \multicolumn{1}{|c}{50.38} & 53.74 & 46.15 & 48.52 & \multicolumn{1}{|c}{54.26} & 36.57 & \multicolumn{1}{|c}{48.27} \\
            \multirow{1}{*}{DeepResearcher-7B} & -- & \ding{55} & \multicolumn{1}{|c}{58.36} & 62.87 & 55.58 & 47.17 & \multicolumn{1}{|c}{52.75} & 40.82 & \multicolumn{1}{|c}{52.93} \\
            \midrule
            \multicolumn{10}{c}{\cellcolor[HTML]{EFEFEF}\textbf{\textit{Open-Sourced Base / Instruct Models}}} \\
            \midrule
            \multirow{6}{*}{Qwen2.5-7B-Instruct} & Naive Response & \ding{55} & \multicolumn{1}{|c}{48.78} & 35.40 & 30.35 & 34.72 & \multicolumn{1}{|c}{40.39} & 33.80 & \multicolumn{1}{|c}{37.24} \\
             & ReAct & \ding{55} & \multicolumn{1}{|c}{52.15} & 37.57 & 34.46 & 40.43 & \multicolumn{1}{|c}{44.08} & 37.52 & \multicolumn{1}{|c}{41.04} \\
             & + MPO & \ding{52} & \multicolumn{1}{|c}{67.31} & 58.64 & 52.28 & 56.76 & \multicolumn{1}{|c}{53.85} & 49.67 & \multicolumn{1}{|c}{56.42} \\
             & SFT & \ding{52} & \multicolumn{1}{|c}{\underline{67.53}} & 63.35 & \underline{73.10} & \underline{74.64} & \multicolumn{1}{|c}{55.68} & 46.92 & \multicolumn{1}{|c}{63.54} \\
             & Vanilla RL & \ding{55} & \multicolumn{1}{|c}{65.49} & \underline{64.78} & 70.76 & 71.28 & \multicolumn{1}{|c}{\underline{58.62}} & \underline{50.59} & \multicolumn{1}{|c}{\underline{63.59}} \\
             & \textbf{\ours (ours)} & \ding{52} & \multicolumn{1}{|c}{\textbf{70.80}} & \textbf{67.84} & \textbf{75.37} & \textbf{77.69} & \multicolumn{1}{|c}{\textbf{61.56}} & \textbf{57.93} & \multicolumn{1}{|c}{\textbf{68.53}} \\
             \midrule
             \multirow{6}{*}{LLaMA3.1-8B-Instruct} & Naive Response & \ding{55} & \multicolumn{1}{|c}{35.63} & 38.56 & 38.22 & 36.40 & \multicolumn{1}{|c}{46.17} & 30.64 & \multicolumn{1}{|c}{37.60} \\
             & ReAct & \ding{55} & \multicolumn{1}{|c}{38.48} & 42.94 & 45.83 & 38.56 & \multicolumn{1}{|c}{47.36} & 36.92 & \multicolumn{1}{|c}{41.68} \\
             & + MPO & \ding{52} & \multicolumn{1}{|c}{54.25} & 50.31 & 43.86 & 52.60 & \multicolumn{1}{|c}{58.92} & 45.33 & \multicolumn{1}{|c}{50.88} \\
             & SFT & \ding{52} & \multicolumn{1}{|c}{\underline{74.92}} & \underline{69.84} & 58.42 & \underline{73.55} & \multicolumn{1}{|c}{55.52} & 50.76 & \multicolumn{1}{|c}{\underline{63.84}} \\
             & Vanilla RL & \ding{55} & \multicolumn{1}{|c}{70.68} & 68.13 & \underline{60.57} & 68.80 & \multicolumn{1}{|c}{\underline{59.74}} & \underline{52.05} & \multicolumn{1}{|c}{63.33} \\
             & \textbf{\ours (ours)} & \ding{52} & \multicolumn{1}{|c}{\textbf{78.53}} & \textbf{72.78} & \textbf{64.76} & \textbf{79.61} & \multicolumn{1}{|c}{\textbf{68.24}} & \textbf{58.68} & \multicolumn{1}{|c}{\textbf{70.43}} \\
             \midrule
             \multirow{6}{*}{Qwen3-8B} & Naive Response & \ding{55} & \multicolumn{1}{|c}{54.08} & 42.14 & 36.37 & 34.95 & \multicolumn{1}{|c}{48.46} & 36.53 & \multicolumn{1}{|c}{42.09} \\
             & ReAct & \ding{55} & \multicolumn{1}{|c}{62.56} & 50.58 & 44.62 & 41.60 & \multicolumn{1}{|c}{54.35} & 42.68 & \multicolumn{1}{|c}{49.40} \\
             & + MPO & \ding{52} & \multicolumn{1}{|c}{65.42} & 54.67 & 46.25 & 48.79 & \multicolumn{1}{|c}{56.81} & 39.50 & \multicolumn{1}{|c}{51.91} \\
             & SFT & \ding{52} & \multicolumn{1}{|c}{64.73} & 62.75 & 63.16 & 75.83 & \multicolumn{1}{|c}{59.67} & 49.25 & \multicolumn{1}{|c}{62.57} \\
             & Vanilla RL & \ding{55} & \multicolumn{1}{|c}{\underline{68.47}} & \textbf{70.29} & \underline{67.35} & \underline{80.42} & \multicolumn{1}{|c}{\underline{63.44}} & \underline{52.04} & \multicolumn{1}{|c}{\underline{67.00}} \\
             & \textbf{\ours (ours)} & \ding{52} & \multicolumn{1}{|c}{\textbf{72.51}} & \underline{69.06} & \textbf{71.48} & \textbf{83.65} & \multicolumn{1}{|c}{\textbf{65.28}} & \textbf{56.62} & \multicolumn{1}{|c}{\textbf{69.77}} \\
            \bottomrule
        \end{tabular}
    }
    \caption{\label{tab:main_results} 
    Comparison of \ours with baselines. 
    ``w/o Plan.'' indicates whether the inference paradigm includes global planning as a mechanism for 
    guidance. 
    The best and second best of each model are in \textbf{bold} and \underline{underlined}.}
    \vspace{-1em}
\end{table*} 

\textbf{Datasets.} \quad 
We collect data from the training splits of four datasets during \textit{training}: ALFWorld~\citep{shridhar2021alfworld}, IQA~\citep{gordon2018iqa}, TextCraft~\citep{prasad-etal-2024-adapt}, and Wordle~\citep{abdulhai2023lmrlwordle}. 
Our \textit{evaluation} is conducted on six benchmarks. 
We employ the test splits of ALFWorld, IQA, TextCraft, and Wordle for in-domain (ID) assessment, and the full dataset samples of MAZE~\citep{abdulhai2023lmrlwordle} and BabyAI~\citep{chevalier-boisvert2018babyai} for out-of-domain (OOD) scenarios. 
We collected data from prior work~\citep{song2024agentbank,xi2024agentgym}, and use only the task instructions and their corresponding final answers for RL-related training, 
with the overall dataset statistics described in 
\Cref{tab:dataset_statistics} and Appendix \ref{subsec:appendix_datasets}. 
We adopt the unified \textit{LLM-as-Judge}~\citep{zheng2023judging,gu2024survey} 
paradigm based on DeepSeek-V3~\citep{liu2024deepseek} 
to verify the model's end-to-end (E2E) performance in a fair comparison protocol, including \textbf{\textit{(1)}} the task completion rates, and \textbf{\textit{(2)}} the efficiency of interaction trajectories, and then calculate the average scores as the evaluation metric.

\textbf{Models and Implementation.} \quad 
We validate the effectiveness of \ours across different base and instruction-tuned models, including 
Qwen2.5-7B-Instruct~\citep{yang2024qwen2}, LLaMA3.1-8B-Instruct~\citep{dubey2024llama}, and Qwen3-8B~\citep{yang2025qwen3}. 
The \textit{reinforcement learning (RL)} framework is built on verl~\citep{sheng2025hybridflow} with Group Relative Policy Optimization (GRPO)~\citep{shao2024deepseekmath} as the learning algorithm. 
The total training dataset contains 5725 samples.
Each sample undergoes 16 rollouts, with a training batch size of 256 and a rollout batch size of 64. The total number of training epochs is set to 4, with 1 epoch allocated to Stage 1, 2 epochs to Stage 2, and an additional 1 epoch dedicated to Stage 3. 
The learning rate is set at 1e\mbox{-}6. 
Following the approach proposed by ~\citet{sun2025zerosearch}, we employ the frontier model DeepSeek-V3 to simulate real-world environmental behaviors. 
Notably, in our training setup, the environmental observation $\mathcal{O}$ is concatenated into the interaction process, which are not generated by the training policy. 
To prevent these segments from influencing gradient updates, we apply masking during loss calculation, where we mask out all content enclosed within \texttt{<observation>...</observation>} tags. 
When conducting \textit{supervised fine-tuning (SFT)} as baseline competitors, we utilized a learning rate scheduler featuring linear warm-up and cosine decay, peaking at a learning rate of 2e\mbox{-}5, alongside a warmup ratio of 0.03 and a weight decay of 0.0 and a batch size of 256 for 4 epochs.

\textbf{Baselines.} \quad 
We compare \ours with the following baselines: 
(1) We employ GPT-4o and GPT-4o-mini~\citep{hurst2024gpt} as the \textbf{\textit{Close-Sourced Models}} competitors. 
(2) \textbf{\textit{Open-Sourced Agent-Specific Models}} include Agent-FLAN-7B~\citep{chen2024agent}, LLaMA-xLAM-2-8B-fc-r~\citep{zhang2024agentohana} and DeepResearcher-7B~\citep{zheng2025deepresearcher}. 
(3) The simplest baseline is \textbf{\textit{Naive Response}}, where the model generates responses directly without any training or prompting strategies. 
(4) \textbf{\textit{ReAct}}~\citep{yao2023react} is the common agent paradigm that prompts agents to integrate single-step reasoning with immediate action execution. 
(5) \textbf{\textit{MPO}}~\citep{xiong2025mpo} acts as an external plug-and-play planner that endows the model with meta-plans to provide explicit guidance during task execution. 
(6) We also perform \textbf{\textit{Supervised Fine-Tuning (SFT)}} on models, a widely adopted training strategy in a series of previous works~\citep{chen2024agent,song2024agentbank,xi2024agentgym,zeng2024agenttuning,zhang2024agent,fuagentrefine}. 
Specifically, we utilize frontier models (e.g., DeepSeek-V3) to generate global plans that guide the execution of target tasks.
(7) \textbf{\textit{Vanilla RL}} is the naive reinforcement learning process that utilizes the Group Relative Policy Optimization (GRPO)~\citep{shao2024deepseekmath} algorithm. In this setup, we utilize only the format and end-to-end (E2E) performance as the reward metrics. 
Details are in Appendix \ref{subsec:appendix_baselines}.

\subsection{Main Results}

The main results of 
\ours are demonstrated in \Cref{tab:main_results}, and 
we summarize the observations below. 

\textbf{\ours is effective across different models.} 
Experimental results in \Cref{tab:main_results} show that our \ours consistently outperforms other baseline approaches on both base and instruction-tuned models in terms of agent task completion. 
Compared to the \textit{naive response}, \ours enhances the average downstream task performances by 78.51\%. 
Remarkably, when compared to \textit{open-sourced agent-specific models} such as DeepResearcher-7B, our approach achieves over 29.47\% higher performance with the same backbone model of Qwen2.5-7B-Instruct. 
In comparison to the plug-and-play external planner \textit{MPO}, our method achieves an average improvement of 31.10\%, further highlighting the importance of tight coordination between the planner and executor in effectively solving agent-oriented tasks. 
Furthermore, open-sourced models enhanced with \ours demonstrate the potential to outperform \textit{close-sourced proprietary models} in agent problem-solving. 
Specifically, models integrated with \ours achieve an average improvement of 2.35\% over GPT-4o, while showing a more substantial gain of 53.90\% over GPT-4o-mini at a comparable parameter scale.

\textbf{AdaPlan paradigm + RL boosts agent performances.} 
Here we focus on analyzing the performance of two baseline methods: \textit{SFT} and \textit{Vanilla RL}. 
The primary distinction between SFT and \ours lies in the training strategies, while the key difference between Vanilla RL and our method is whether to incorporate the AdaPlan paradigm to provide global guidance for agent execution. 
As presented in \Cref{tab:main_results}, the average performance of SFT and Vanilla RL is quite similar on both Qwen2.5-7B-Instruct and LLaMA3.1-8B-Instruct. 
This suggests that the enhancement brought by global plan guidance in SFT is roughly on par with the incremental gain achieved through RL-based training. 
Specifically, for in-domain (ID) tasks, SFT outperforms Vanilla RL by a marginal average of 2.75\%, whereas Vanilla RL achieves an average lead of 5.80\% in out-of-domain (OOD) tasks. 
For reasoning-oriented models such as Qwen3-8B, which inherently possess a certain degree of multi-step reasoning and decision-making capabilities required for complex agent tasks, the performance gains from the AdaPlan paradigm are insufficient to offset the advantages of RL over SFT training. 
In contrast, PilotRL demonstrates robust performance gains across models with diverse characteristics, achieving consistent improvements over both SFT and Vanilla RL by 9.89\% and 7.64\%, respectively. 
These observations further highlight the importance of combining the global planning capabilities of the AdaPlan paradigm with RL training, as embodied in our \ours framework, for enhancing model performance in complex agent scenarios.

\section{Ablations and Analysis}
\label{sec:ablation_study}

We conduct ablation studies to highlight the contribution of each training stage and the impact of their sequential order on PilotRL. 
Furthermore, we perform an in-depth analysis of PilotRL's effectiveness, examining key aspects such as our AdaPlan paradigm, 
the architecture of unified planner-executor, 
and the co-evolution of components. 

\subsection{Necessity of Progressive Training} 

\begin{table}[tb] 
    \vspace{-4mm}
    \centering
    \setlength{\tabcolsep}{2pt}
    \resizebox{1\linewidth}{!}{
    \begin{tabular}{c|c|c|c}
        \toprule
        Order & In-Domain & Out-of-Domain & Avg. \\ 
        \midrule
        \multicolumn{4}{c}{\cellcolor[HTML]{EFEFEF}\textit{Standard Pipeline}}\\
        \midrule
        \multirow{1}*{\textbf{1 $\rightarrow$ 2 $\rightarrow$ 3}} 
        & \textbf{73.68} & \textbf{61.39} & \textbf{69.58} \\
        \midrule
        \multicolumn{4}{c}{\cellcolor[HTML]{EFEFEF}\textit{Necessity of Progressive Training}}\\
        \midrule
        \multirow{1}*{1 $\&$ 2 $\&$ 3} 
      & 71.64 ($\downarrow$ 2.77\%) & 58.52 ($\downarrow$ 4.68\%) & 67.27 ($\downarrow$ 3.32\%) \\
        \midrule
        \multicolumn{4}{c}{\cellcolor[HTML]{EFEFEF}\textit{The Role of Each Stage}}\\
        \midrule
        \multirow{1}*{2 $\rightarrow$ 3} 
        & 70.82 ($\downarrow$ 3.88\%) & 58.33 ($\downarrow$ 4.98\%) & 66.66 ($\downarrow$ 4.20\%) \\
        \multirow{1}*{1 $\rightarrow$ 3} 
        & 70.66 ($\downarrow$ 4.10\%) & 58.39 ($\downarrow$ 4.89\%) & 66.57 ($\downarrow$ 4.33\%) \\
        \multirow{1}*{1 $\rightarrow$ 2} 
        & 72.21 ($\downarrow$ 2.00\%) & 59.02 ($\downarrow$ 3.86\%) & 67.81 ($\downarrow$ 2.54\%) \\
        \midrule
        \multicolumn{4}{c}{\cellcolor[HTML]{EFEFEF}\textit{Sequential Order of Stages}}\\
        \midrule
        \multirow{1}*{2 $\rightarrow$ 1 $\rightarrow$ 3} 
        & \underline{72.79} ($\downarrow$ 1.21\%) & \underline{59.88} ($\downarrow$ 2.46\%) & \underline{68.48} ($\downarrow$ 1.58\%) \\
        \bottomrule
    \end{tabular}
    }
    \caption{\label{tab:ablation}
    Analysis of the training stages and sequential order. 
    ``Order'' refers to the sequence of Stage 1, 2, and 3. 
    ``1 $\&$ 2 $\&$ 3'' denotes a training setting in which the reward functions from all stages are applied simultaneously. 
    We compute the \textit{average} performance of the evaluated models across each benchmark. 
    The best and second best scores are in \textbf{bold} and \underline{underlined}.
    }
    \vspace{-1em}
\end{table}

We aggregated the reward functions from all training stages to verify the importance of incrementally optimizing the planning and execution capabilities in a staged and progressive manner. 
Results are presented in \Cref{tab:ablation} (\underline{1 $\&$ 2 $\&$ 3}), where we observe a performance drop of 3.32\% compared to our multi-stage training strategy (\underline{1 $\rightarrow$ 2 $\rightarrow$ 3}). 
A primary cause of this performance drop lies in the intrinsic complexity and potential conflicts among heterogeneous reward signals. Specifically, the planning-oriented and execution-driven components exert distinct behavioral demands on the model, which can lead to unstable policy updates during training. For instance, 
early in training, 
the model may lack a sufficiently mature structure for guidance follow-up, making it difficult to accurately adhere to global plans. It results in conflicting gradient signals and ultimately reduces learning efficiency.

\subsection{The Role of Each Stage}
\label{subsubsec:training_stage}

To assess the contribution of each individual stage, we conduct three ablation studies by sequentially removing Stage 1, 2, and 3, respectively. The models are then evaluated on both in-domain (ID) and out-of-domain (OOD) benchmark tasks, with the results presented in \Cref{tab:ablation}. 
To ensure a fair comparison and control for the impact of training data volume on performance, 
we fix the total number of training epochs at 4, which is consistent with the main experimental setup, and allocate 2 epochs to each of the remaining two stages for training. 
Detailed analysis is provided in Appendix \ref{subsubsec:appendix_training_stage}.

\subsection{Sequential Order of Stages}

We swap Stage 1 and Stage 2 to evaluate their influence on model performance. As seen in \Cref{tab:ablation} (\underline{2 $\rightarrow$ 1 $\rightarrow$ 3}), such reordering results in a slight performance decline of 1.58\%. 
It supports the robustness of our original training sequence, 
which prioritizes the development of guidance-following capabilities before refining plan generation skills. 
It is grounded in the need for a strong foundation of instruction follow-up to enhance the quality of global plans. Only with this foundation can the model make meaningful strides in developing its ability to generate global plans that effectively guide the action execution during agent task completion.

\subsection{Further Analysis}

\begin{table}[tb] 
    \vspace{-4mm}
    \centering
    \resizebox{1\linewidth}{!}{
    \begin{tabular}{c|c|c|c|c}
    \toprule
    Backbone Model & Paradigm & ID & OOD & Avg. \\ 
    \midrule
    \multirow{2}*{\makecell{Qwen2.5-7B\\-Instruct}} & AdaPlan & \textbf{50.54} & \textbf{44.98} & \textbf{48.69} \\
    & ReAct & 41.15 & 40.80 & 41.04 \\
    \midrule
    \multirow{2}*{\makecell{LLaMA3.1-8B\\-Instruct}} & AdaPlan & \textbf{47.42} & \textbf{47.20} & \textbf{47.34} \\
    & ReAct & 41.45 & 42.14 & 41.68 \\
    \midrule
    \multirow{2}*{Qwen3-8B} & AdaPlan & \textbf{53.69} & \textbf{51.49} & \textbf{52.95} \\
    & ReAct & 49.84 & 48.52 & 49.40 \\
    \bottomrule
    \end{tabular}
    }
    \caption{\label{tab:paradigm_ablation}
    Analysis on the agent paradigms of \textit{AdaPlan} and \textit{ReAct} on In-Domain (ID) and Out-of-Domain (OOD) tasks. The best score of each model are in \textbf{bold}.
    }
\end{table}

\textbf{AdaPlan vs. ReAct.} 
We compare the performance of the AdaPlan and ReAct paradigms in agent tasks. 
Neither of these paradigms undergoes additional training, with distinct prompt strategies employed to induce different thinking patterns in the model instead. 
As presented in \Cref{tab:paradigm_ablation}, the results indicate that our proposed AdaPlan exhibits greater efficacy in enabling the model to accomplish complex agent tasks by leveraging global planning as guidance, which outperforms ReAct by 12.76\%.

\begin{table}[tb] 
    \centering
    \resizebox{1\linewidth}{!}{
    \begin{tabular}{c|c|c|c|c}
    \toprule
    Backbone Model & Architecture & ID & OOD & Avg. \\ 
    \midrule
    \multirow{2}*{\makecell{Qwen2.5-7B\\-Instruct}} & Unified & \textbf{72.93} & \textbf{59.75} & \textbf{68.53} \\
    & Isolated & 68.94 & 55.18 & 64.36 \\
    \midrule
    \multirow{2}*{\makecell{LLaMA3.1-8B\\-Instruct}} & Unified & \textbf{73.92} & \textbf{63.46} & \textbf{70.43} \\
    & Isolated & 68.68 & 59.18 & 65.51 \\
    \midrule
    \multirow{2}*{Qwen3-8B} & Unified & \textbf{74.18} & \textbf{60.95} & \textbf{69.77} \\
    & Isolated & 72.66 & 56.02 & 67.11 \\
    \bottomrule
    \end{tabular}
    }
    \caption{\label{tab:coordination_ablation}
    Analysis of the \textit{unified} and \textit{isolated} planner-executor architectures on In-Domain (ID) and Out-of-Domain (OOD) tasks. The best scores are in \textbf{bold}.
    }
    \vspace{-1em}
\end{table}

\textbf{Unified Architecture vs. Isolated Planner-Executor Architecture.} 
We conduct an evaluation against the isolated planner and executor framework~\citep{erdogan2025plan} to validate the effectiveness of integrating both components within a unified model architecture. 
In the isolated architecture setting, we employ the same backbone model and separately train the planner and executor modules following the Stage 1 and Stage 2 RL procedures described in PilotRL, with each component trained for 2 epochs. 
As summarized in \Cref{tab:coordination_ablation}, the isolated architecture suffers from a performance drop of 5.63\% compared to the unified architecture, in which both functionalities are learned jointly in an end-to-end manner, 
further emphasizing the importance of co-developing planning and execution capabilities within a single model.

\begin{figure}[t]
    \centering
    \includegraphics[width=1\linewidth]{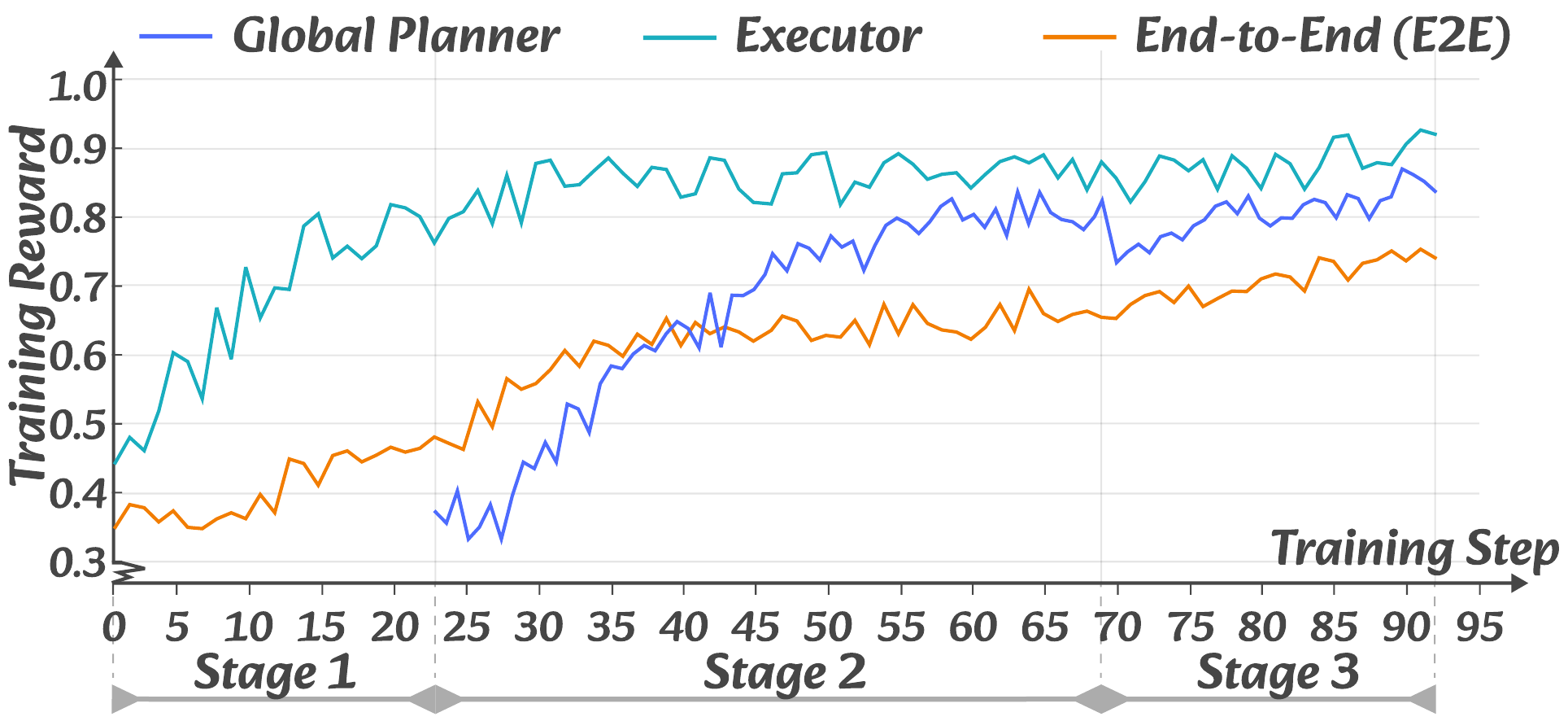}
    \vspace{-1em}
    \caption{Normalized rewards for global planner, executor and end-to-end (E2E) performance in training LLaMA3.1-8B-Instruct.}
    \label{fig:line_chart}
    \vspace{-1em}
\end{figure}

\textbf{How planner, executor, and their coordination co-evolve during agent learning?} 
We analyze the evolution of reward scores for the global planner, the executor, and the end-to-end (E2E) performance in the training process of LLaMA3.1-8B-Instruct. 
As shown in \Cref{fig:line_chart}, the executor’s ability of plan adhesion 
saw a marked improvement during Stage 1 and remained stable with slight growth in subsequent stages.
The global planner's performance, which generates high-level plans for explicit guidance, exhibits a notable improvement in Stage 2 (epoch 2 \& 3). It experiences a mild decline at the beginning of Stage 3, followed by a continuous upward trend. We speculate that this temporary drop reflects an adaptation period, during which the planner adjusts its generation to better align with the executor’s capabilities.
Meanwhile, the E2E reward increases steadily throughout the entire training process, indicating a consistent improvement in the system’s overall performance. 

\section{Related Work}
\label{sec:related_work}

\textbf{LLM as Agent} \quad 
The emergence of Large Language Models (LLMs) has driven research into the development of LLM-based agent systems~\citep{zeng2024agenttuning}. 
The most common paradigm 
is ReAct~\citep{yao2023react}, which integrates Chain-of-Thought (CoT) reasoning with action in an interleaved manner to accomplish 
tasks. 
However, this step-by-step reasoning framework struggles in scenarios demanding 
complex multi-step coordination, 
e.g., household exploration~\citep{shridhar2021alfworld} and games involving foresighted planning~\citep{abdulhai2023lmrlwordle},  
which highlights a pressing need for long-term planning. 
Even though there have been efforts aimed to incorporate explicit guidance into agent task completion~\citep{deng2023mind2web,zeng2024agenttuning}, the planner and executor are typically implemented in isolated architectures, leading to suboptimal guidance generation and execution alignment. 
Moreover, although closed-source models often demonstrate strong performance in agent tasks, open-source models generally fall short in comparison~\citep{liu2023agentbench}. 
While studies have tried to collect expert trajectories from frontier LLMs (e.g., GPT-4) to fine-tune open-sourced models~\citep{chen2023fireact,chen2024agent,song2024agentbank,zeng2024agenttuning,zhang2024agent}, such behavioral cloning strategy hinders the model's generalizability on out-of-distribution tasks. 
Therefore, it is necessary to introduce a more flexible training framework to cultivate models' intrinsic generalization capabilities, e.g., reinforcement learning. 

\textbf{Reinforcement Learning in LLMs} \quad 
Compared to the supervised fine-tuning (SFT), reinforcement learning (RL) provides a more powerful paradigm for training LLM-based agents which are capable of decision-making without explicit supervision~\citep{guo2025deepseek,jaech2024openai,team2025kimi}. 
Among all the RL algorithms, GRPO~\citep{shao2024deepseekmath,guo2025deepseek} is specifically designed for LLMs, which has proven to be highly effective by replacing the traditional 
critic with a group-based evaluation strategy.  
Efforts have been made to enhance the agent capability in LLMs through the RL process, with notable works for information retrieval tasks~\citep{jin2025search,song2025r1} and tool utilization scenarios~\citep{feng2025retool,li2025torl}. 
We situate our research on agent capability enhancement within the RL landscape for its effectiveness in fostering exploration and the emergence of novel strategies, and shift away from the commonly used ReAct framework~\citep{yao2023react}, toward a global-plan-driven paradigm that supports more strategic and forward-looking decision-making.

\section{Conclusion}
\label{sec:conclusion}

In this paper, we introduce AdaPlan, an adaptive global plan-based agent paradigm. Based on the proposed paradigm, we put forward PilotRL, a global planning-guided training framework for LLM agents driven by progressive reinforcement learning. 
Experimental results indicate that \ours achieves excellent outcomes in agent scenarios. 

\section{Limitations}
\label{sec:limitations}

There are some limitations in our work. 
The generation of initial global plans, as well as the evaluation of model performance during the training process, relies on advanced large language models. This introduces a dependency and may lead to the propagation of biases.

\section{Ethical Considerations}
\label{sec:ethical_considerations}

The experimental design in our paper was carefully planned to ensure that all data used for training and evaluation were obtained through legitimate means and adhered to relevant privacy laws and regulations. 
We have also provide detailed descriptions of our methodologies, algorithms, and prompts to enable reproducibility.


\bibliography{PilotRL_2026}

\clearpage
\clearpage


\appendix

\renewcommand{\contentsname}{Appendix}
\tableofcontents
\addtocontents{toc}{\protect\setcounter{tocdepth}{2}}



\section{Group Relative Policy Optimization (GRPO)}

We utilize the Group Relative Policy Optimization (GRPO) as the RL algorithm. 
For each question $x \sim \mathcal{D}$, the behavior policy $\pi_{\theta_{\text{old}}}$ generates a set of $G$ candidate completions $\tau = \{y_i\}_{i=1}^{G} \sim \pi_{\theta_{\text{old}}}(\cdot|x)$,  
with each response receiving a scalar reward $r_i$. 
The training objective is to optimize the policy $\pi_{\theta}$ based on reference policy $\pi_{\theta_{\text{ref}}}$: 
\begin{equation}
\resizebox{1\linewidth}{!}{
$\begin{split}
  \mathcal{J}(\theta) & = \mathbb{E}_{x \sim \mathcal{D}, \{y_i\}_{i=1}^{G} \sim \pi_{\theta_{\text{old}}}(\cdot|x)} \frac{1}{G} \sum_{i=1}^{G} [  
  \min ( \frac{\pi_{\theta}(y_i|x)}{\pi_{\theta_{\text{old}}}(y_i|x)} \hat{A_{i}}, \\ 
  & \text{clip} ( \frac{\pi_{\theta}(y_i|x)}{\pi_{\theta_{\text{old}}}(y_i|x)}, 1-\epsilon, 1+\epsilon ) \hat{A_{i}} ) - \beta \mathbb{D}_{\text{KL}} ( \pi_{\theta} || \pi_{\theta_{\text{ref}}} ) ]
\end{split}$
}
\label{eq:grpo}
\end{equation}
where the group-normalized advantage $\hat{A_{i}}$ of the $i$-th rollout in current group is defined as: 
\begin{equation*}
    \hat{A_{i}} = \frac{r_i - \text{mean}(\{r_j\}_{j=1}^{G})}{\text{std}(\{r_j\}_{j=1}^{G})}
    \label{eq:group_normalized_advantage}
\end{equation*}

\begin{figure*}[ht!]
    \centering
    \includegraphics[width=1\linewidth]{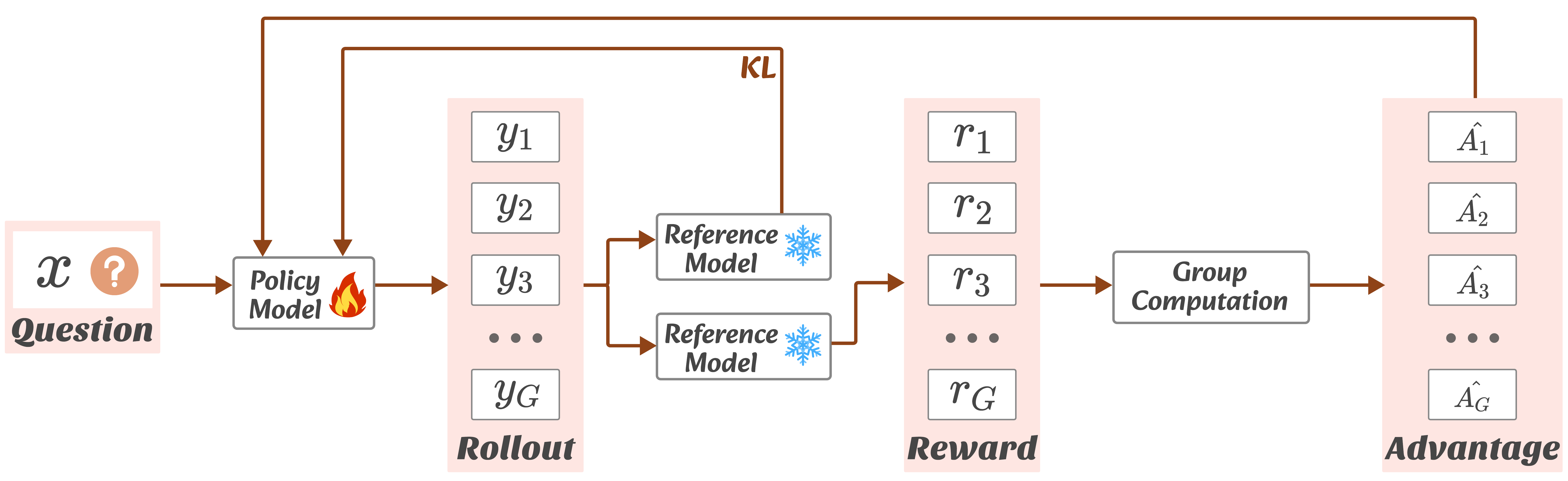}
    \caption{An illustration for the Group Relative Policy Optimization (GRPO) pipeline.}
    \label{fig:GRPO}
\end{figure*}

An overview of the GRPO algorithm is illustrated in \Cref{fig:GRPO}. In this formulation, $\epsilon$ denotes the clipping ratio, a hyperparameter that controls the allowable deviation between the updated and reference policies. The \texttt{clip} function restricts the importance weight $r_i$ within the range $\left[1-\epsilon, 1+\epsilon\right]$, which enhances training stability and reduces the risk of policy collapse. 
The parameter $\beta$ represents the Kullback–Leibler (KL) loss coefficient~\citep{hall1987kullback}, which governs the strength of the KL divergence penalty included in the objective function. This penalty term helps constrain the policy updates, ensuring that the learned policy remains sufficiently close to the original reference policy and thereby improving overall training stability. 

\section{Experiment Details}

\subsection{Datasets}
\label{subsec:appendix_datasets}

To evaluate the performance of \ours, we conduct experiments using six datasets for agent tasks. Specifically, four datasets are used for training and in-domain (ID) performance evaluation, while the remaining two are reserved for out-of-domain (OOD) assessment, as shown in \Cref{tab:dataset_statistics}.

\begin{itemize}[leftmargin=*]
    \item \textbf{ALFWorld}~\citep{shridhar2021alfworld}: It is a home-oriented environment built upon TextWorld, where agents are required to navigate through rooms and apply common sense reasoning to perform various tasks. It mirrors the embodied settings found in the ALFRED dataset~\citep{shridhar2020alfred}, and offers human-annotated ideal trajectories for use in imitation learning. 
    \item \textbf{IQA}~\citep{gordon2018iqa}: The Interactive QA dataset is a question answering task in which an agent need to engage with a dynamic visual environment to find answers. Here we utilize the text version from ~\citet{jia-etal-2024-langsuit}. 
    \item \textbf{TextCraft}~\citep{prasad-etal-2024-adapt}: It is a text-only environment for crafting Minecraft items that resembles cooking recipes with steps of varying complexity. This dataset  exhibits an inherently decomposable structure, providing a more suitable environment for our proposed paradigm. 
    \item \textbf{Wordle}~\citep{abdulhai2023lmrlwordle}: It is a word-guessing game designed to assess agents' reasoning capabilities at the letter level, where the agents attempt to identify a target word selected from a predefined vocabulary consisting of five-letter words. In order to successfully identify the target word with minimum trials within the limited number of allowed attempts, it is crucial for the model to employ efficient global planning. 
    \item \textbf{MAZE}~\citep{abdulhai2023lmrlwordle}: This dataset is also a word-based puzzle game in which agents, serving as players, are aware of their current position, the location of the goal, and the presence of walls in the four cardinal directions, e.g., up, down, left, and right. 
    \item \textbf{BabyAI}~\citep{chevalier-boisvert2018babyai}: The BabyAI dataset evaluates agent performance in embodied navigation and interaction scenarios. It features a simulated grid-world environment containing 40 instruction-following tasks, where agents are required to understand commands and interact with objects accordingly.
\end{itemize}

We have collected the data for training and evaluation from \citet{song2024agentbank} and \citet{xi2024agentgym}. For the ALFWorld and IQA data, we utilize the datasets as provided in ~\citet{song2024agentbank}, while for TextCraft, Wordle, MAZE, and BabyAI, we adopt the versions from ~\citet{xi2024agentgym}. 
The reference trajectories included in these original data sources are used exclusively for supervised fine-tuning (SFT) of the baselines. During both the reinforcement learning (RL) training and evaluation phases, we only make use of the task instructions and their corresponding final answers.

\begin{table}[t]
    \small
    \centering
    \vspace{0.3cm}
    \setlength{\tabcolsep}{4pt}
    \resizebox{\columnwidth}{!}{
        \begin{tabular}{c|c|ccc}
            \toprule
            Classification & Dataset & \#Training Num. & \#Testing Num. \\
            \midrule
            \multirow{4}{*}{\textit{In-Domain}} & ALFWorld & 3000 & 321 \\
             & IQA & 1465 & 162 \\
             & TextCraft & 400 & 74 \\
             & Wordle & 860 & 95 \\
            \midrule
             \multirow{2}{*}{\textit{Out-of-Domain}} & BabyAI & -- & 400 \\
             & MAZE & -- & 215 \\
            \bottomrule
        \end{tabular}
    }
    \caption{\label{tab:dataset_statistics}
        Statistics of data for training and evaluation.}
\end{table}

\subsection{Baselines}
\label{subsec:appendix_baselines}

In this section, we provide a comprehensive overview of the various methods that serve as baselines in our comparison.

\begin{itemize}[leftmargin=*]
    \item \textbf{Close-Sourced Models}: Closed-source models are considered to represent the current state-of-the-art in LLM capabilities and are regarded as the most competitive baseline methods. We have selected GPT-4o and GPT-4o-mini~\citep{hurst2024gpt} to assess the upper bound of the model performance on agent tasks. 
    \item \textbf{Open-Sourced Agent-Specific Models}: These models refer to models that were trained specifically on agent-task datasets. We have selected Agent-FLAN-7B~\citep{chen2024agent}, LLaMA-xLAM-2-8B-fc-r~\citep{zhang2024agentohana} and DeepResearcher-7B~\citep{zheng2025deepresearcher} to represent the open-sourced agent-specific models for comparison to assess PilotRL's relative advantages. Specifically, the backbone model of DeepResearcher-7B is Qwen2.5-7B-Instruct~\citep{yang2024qwen2}, which facilitates a more direct comparison with Qwen2.5-7B-Instruct + PilotRL. 
    \item \textbf{Naive Response}: It refers to the case where the model directly generates responses without any training (e.g., SFT, RL, etc.) or prompting (e.g., ReAct) strategies. 
    \item \textbf{ReAct}~\citep{yao2023react}: It is the prompting strategy that integrates single-step reasoning with the execution of the current action, which is a common agent paradigm. 
    \item \textbf{MPO}~\citep{xiong2025mpo}: 
    The Meta Plan Optimization (MPO) framework improves the agent's planning capabilities by integrating explicit guidance into the decision-making process. As an external plug-and-play planner, MPO provides the model with high-level meta-plans that serve as structured guidance during task execution. 
    One key distinction between MPO and \ours lies in the integration and training of the planner and executor components. In our approach, both planner and executor reside within the same model and are trained jointly. In contrast, MPO maintains separate models for planning and execution, where only the planner is trained while the executor's parameters remain frozen, leading to limited coordination between the two components. 
    \item \textbf{Supervised Fine-Tuning (SFT)}: This training strategy is widely adopted in a series of studies~\citep{chen2024agent,song2024agentbank,xi2024agentgym,zeng2024agenttuning,zhang2024agent,fuagentrefine}. 
    However, existing studies have shown that compared to RL, SFT generally exhibits weaker generalization capabilities on new tasks—particularly when the training data consists of multi-step trajectories for problem-solving~\citep{shao2024deepseekmath,team2025kimi}. This is because such trajectories may contain redundant or suboptimal paths to task completion. Moreover, SFT tends to bias the model toward previously seen execution paths, limiting its ability to adapt or generalize to novel scenarios through compositional or analogical reasoning. 
    During SFT, we use the same datasets with PilotRL. In addition, we incorporate the original agent-environment interaction trajectories into training, a setting that differs from \textbf{Vanilla RL} and our \textbf{PilotRL}. 
    Furthermore, we generate global plans for guiding task completion using DeepSeek-V3, and feed both the interaction trajectories and the corresponding global plans into the model during training. 
    This setup allows us to compare \ours over existing baselines under a more fair and controlled experimental condition. 
    \item \textbf{Vanilla RL}: We also conduct training with the naive reinforcement learning process utilizing the Group Relative Policy Optimization (GRPO)~\citep{shao2024deepseekmath} algorithm. Here we employ only the format and end-to-end (E2E) performance as the reward metrics. This baseline is for validating the effectiveness of adaptive global planning. 
\end{itemize}

We ensured a \textit{fair comparison protocol} across all baselines through the following strategies:

\begin{itemize}
    \item \textit{Unified Evaluation Mechanism}: For all baselines, we adopted a consistent LLM-as-Judge paradigm to evaluate their E2E performances with the same \hyperlink{E2E}{prompt} described in Appendix \ref{appendix:prompts}. 
    In our main experiments, DeepSeek-V3 served as the judge model for this paradigm.
    \item \textit{Reward Modeling}: RL-related approaches involve a reward model during training. Among our compared baselines, only Vanilla RL utilizes reinforcement learning (RL). 
    Methods like ReAct (a prompting strategy) and MPO (an external plug-and-play planner) do not rely on reward signals from frontier models.
    For Vanilla RL, we employed format and end-to-end (E2E) performance as the reward metrics. The criteria for these rewards were implemented identically to our PilotRL, which were uniformly applied using DeepSeek-V3 as the frontier reward model with prompts for judgment described in Appendix \ref{appendix:prompts}.
\end{itemize}

\begin{table*}[t] \centering
    \resizebox{1\linewidth}{!}{
        \begin{tabular}{c|cccccccc}
            \toprule
                \multirow{2}{*}{Order} & \multirow{2}{*}{Backbone Model} & \multicolumn{1}{|c}{ALFWorld} & IQA & TextCraft & Wordle & \multicolumn{1}{|c}{BabyAI} & \multicolumn{1}{c|}{MAZE} & \multirow{2}{*}{\textbf{Avg.}} \\
            \cline{3-8}
             & & \multicolumn{4}{|c}{\textit{In-Domain (ID)}} &  \multicolumn{2}{|c|}{\textit{Out-of-Domain (OOD)}} &  \\
            \midrule
            \multicolumn{9}{c}{\cellcolor[HTML]{EFEFEF}\textit{\textbf{Standard Pipeline}}}\\
            \midrule
            \multirow{3}*{\makecell*[c]{\shortstack{\textbf{1 $\rightarrow$ 2 $\rightarrow$ 3}} \\ \textbf{(ours)}}} 
            & Qwen2.5-7B-Instruct & \multicolumn{1}{|c}{\textbf{70.80}} & \textbf{67.84} & \underline{75.37} & \underline{77.69} & \multicolumn{1}{|c}{\textbf{61.56}} & \textbf{57.93} & \multicolumn{1}{|c}{\textbf{68.53}} \\
            & LLaMA3.1-8B-Instruct & \multicolumn{1}{|c}{\textbf{78.53}} & \underline{72.78} & \textbf{64.76} & \textbf{79.61} & \multicolumn{1}{|c}{\textbf{68.24}} & \textbf{58.68} & \multicolumn{1}{|c}{\textbf{70.43}} \\
            & Qwen3-8B & \multicolumn{1}{|c}{\underline{72.51}} & 69.06 & \textbf{71.48} & \textbf{83.65} & \multicolumn{1}{|c}{\textbf{65.28}} & \textbf{56.62} & \multicolumn{1}{|c}{\textbf{69.77}} \\
            \midrule
            \multicolumn{9}{c}{\cellcolor[HTML]{EFEFEF}\textit{\textbf{Necessity of Progressive Training}}}\\
            \midrule
            \multirow{3}*{1 $\&$ 2 $\&$ 3} 
            & Qwen2.5-7B-Instruct & \multicolumn{1}{|c}{68.29} & 65.43 & 72.91 & 75.82 & \multicolumn{1}{|c}{57.98} & 54.37 & \multicolumn{1}{|c}{65.80} \\
            & LLaMA3.1-8B-Instruct & \multicolumn{1}{|c}{75.56} & 70.42 & 63.03 & 74.51 & \multicolumn{1}{|c}{63.74} & 56.00 & \multicolumn{1}{|c}{67.21} \\
            & Qwen3-8B & \multicolumn{1}{|c}{70.89} & 71.30 & 69.68 & 81.84 & \multicolumn{1}{|c}{63.19} & \underline{55.81} & \multicolumn{1}{|c}{68.79} \\
            \midrule
            \multicolumn{9}{c}{\cellcolor[HTML]{EFEFEF}\textit{\textbf{Effectiveness of Stage 1  (Instruction Adherence)}}} \\
            \midrule
            \multirow{3}*{2 $\rightarrow$ 3} 
            & Qwen2.5-7B-Instruct & \multicolumn{1}{|c}{66.37} & 63.85 & 72.16 & 74.93 & \multicolumn{1}{|c}{60.05} & 52.54 & \multicolumn{1}{|c}{64.98} \\
            & LLaMA3.1-8B-Instruct & \multicolumn{1}{|c}{73.86} & 70.19 & 63.75 & 72.66 & \multicolumn{1}{|c}{64.37} & 54.93 & \multicolumn{1}{|c}{66.63} \\
            & Qwen3-8B & \multicolumn{1}{|c}{70.97} & 69.63 & 70.12 & 81.35 & \multicolumn{1}{|c}{63.96} & 54.10 & \multicolumn{1}{|c}{68.36} \\
            \midrule
            \multicolumn{9}{c}{\cellcolor[HTML]{EFEFEF}\textit{\textbf{Effectiveness of Stage 2 (Global Planner Cultivation)}}}\\
            \midrule
            \multirow{3}*{1 $\rightarrow$ 3} 
            & Qwen2.5-7B-Instruct & \multicolumn{1}{|c}{66.72} & \underline{66.38} & 71.74 & 76.56 & \multicolumn{1}{|c}{58.85} & 53.48 & \multicolumn{1}{|c}{65.62} \\
            & LLaMA3.1-8B-Instruct & \multicolumn{1}{|c}{73.04} & 72.43 & 61.59 & 70.47 & \multicolumn{1}{|c}{\underline{66.32}} & 53.26 & \multicolumn{1}{|c}{66.19} \\
            & Qwen3-8B & \multicolumn{1}{|c}{70.56} & 68.36 & 69.04 & 80.98 & \multicolumn{1}{|c}{64.47} & 53.95 & \multicolumn{1}{|c}{67.89} \\
            \midrule
            \multicolumn{9}{c}{\cellcolor[HTML]{EFEFEF}\textit{\textbf{Effectiveness of Stage 3 (Dual-Process Collaboration)}}} \\
            \midrule
            \multirow{3}*{1 $\rightarrow$ 2} 
            & Qwen2.5-7B-Instruct & \multicolumn{1}{|c}{67.49} & 65.82 & \textbf{75.65} & 73.34 & \multicolumn{1}{|c}{\underline{60.78}} & 53.17 & \multicolumn{1}{|c}{66.04} \\
            & LLaMA3.1-8B-Instruct & \multicolumn{1}{|c}{75.40} & 71.55 & 62.88 & 75.67 & \multicolumn{1}{|c}{65.19} & 56.92 & \multicolumn{1}{|c}{67.94} \\
            & Qwen3-8B & \multicolumn{1}{|c}{72.18} & \underline{72.61} & \underline{70.59} & \underline{83.27} & \multicolumn{1}{|c}{64.73} & 53.28 & \multicolumn{1}{|c}{\underline{69.44}} \\
            \midrule
            \multicolumn{9}{c}{\cellcolor[HTML]{EFEFEF}\textit{\textbf{Sequential Order of Stages}}}\\
            \midrule
            \multirow{3}*{2 $\rightarrow$ 1 $\rightarrow$ 3} 
            & Qwen2.5-7B-Instruct & \multicolumn{1}{|c}{\underline{70.12}} & 66.08 & 73.98 & \textbf{77.85} & \multicolumn{1}{|c}{59.63} & \underline{55.67} & \multicolumn{1}{|c}{\underline{67.22}} \\
            & LLaMA3.1-8B-Instruct & \multicolumn{1}{|c}{\underline{77.25}} & \textbf{73.15} & \underline{64.02} & \underline{77.63} & \multicolumn{1}{|c}{65.98} & \underline{58.14} & \multicolumn{1}{|c}{\underline{69.36}} \\
            & Qwen3-8B & \multicolumn{1}{|c}{\textbf{72.94}} & \textbf{73.86} & 68.55 & 78.02 & \multicolumn{1}{|c}{\underline{65.07}} & 54.80 & \multicolumn{1}{|c}{68.87} \\
            \bottomrule
        \end{tabular}
    }
    \caption{\label{tab:ablation_detail}
    \textbf{Original scores for each benchmark of the ablation study on multiple training stages and sequential order.} 
    It is the detailed version of \underline{\Cref{tab:ablation} }. 
    ``Order'' is the sequential order of Stage 1, 2, and 3 during training. Specifically, ``1 $\&$ 2 $\&$ 3'' refers to a joint training configuration in which reward functions from all three stages are merged and optimized concurrently, where the target model generates global plans independently throughout the entire training process. 
    The best and second best scores of each model are in \textbf{bold} and \underline{underlined}. }
\end{table*}

\begin{table*}[ht!] 
    \centering
	\resizebox{1\linewidth}{!}{
		\begin{tabular}{c|ccccccc}
            \toprule
                \multirow{2}{*}{Backbone Model} & \multirow{2}{*}{Paradigm} & \multicolumn{1}{|c}{ALFWorld} & IQA & TextCraft & Wordle & \multicolumn{1}{|c}{BabyAI} & \multicolumn{1}{c}{MAZE} \\
            \cline{3-8}
             & & \multicolumn{4}{|c}{\textit{In-Domain (ID)}} &  \multicolumn{2}{|c}{\textit{Out-of-Domain (OOD)}} \\
            \midrule
			\multirow{2}{*}{\makecell{Qwen2.5-7B\\-Instruct}} & ReAct & \multicolumn{1}{|c}{52.15} & 37.57 & 34.46 & 40.43 & \multicolumn{1}{|c}{44.08} & 37.52  \\
            & AdaPlan & \multicolumn{1}{|c}{\textbf{59.72} ($\uparrow$ 14.52\%)} & \textbf{43.68} ($\uparrow$ 16.26\%) & \textbf{45.54} ($\uparrow$ 32.15\%) & \textbf{53.23} ($\uparrow$ 31.66\%) & \multicolumn{1}{|c}{\textbf{47.90} ($\uparrow$ 8.67\%)} & \textbf{42.05} ($\uparrow$ 12.07\%) \\
			\midrule
			\multirow{2}{*}{\makecell{LLaMA3.1-8B\\-Instruct}} & ReAct & \multicolumn{1}{|c}{38.48} & 42.94 & 45.83 & 38.56 & \multicolumn{1}{|c}{47.36} & 36.92 \\
            & AdaPlan & \multicolumn{1}{|c}{\textbf{44.19} ($\uparrow$ 14.84\%)} & \textbf{48.02} ($\uparrow$ 11.83\%) & \textbf{46.67} ($\uparrow$ 1.83\%) & \textbf{50.78} ($\uparrow$ 31.69\%) & \multicolumn{1}{|c}{\textbf{54.46} ($\uparrow$ 14.99\%)} & \textbf{39.94} ($\uparrow$ 8.18\%) \\
            \midrule
            \multirow{2}{*}{Qwen3-8B} & ReAct & \multicolumn{1}{|c}{62.56} & 50.58 & 44.62 & 41.60 & \multicolumn{1}{|c}{54.35} & 42.68 \\
            & AdaPlan & \multicolumn{1}{|c}{\textbf{63.34} ($\uparrow$ 1.25\%)} & \textbf{53.82} ($\uparrow$ 6.41\%) & \textbf{44.98} ($\uparrow$ 0.81\%) & \textbf{52.61} ($\uparrow$ 26.47\%) & \multicolumn{1}{|c}{\textbf{55.73} ($\uparrow$ 2.54\%)} & \textbf{47.24} ($\uparrow$ 10.68\%) \\
			\bottomrule
		\end{tabular}
	}
    \caption{\label{tab:paradigm_ablation_detail}
    \textbf{Original scores for each benchmark of the agent paradigm analysis.} 
    It is the detailed version of \underline{\Cref{tab:paradigm_ablation}}.
    The best scores of each model are in \textbf{bold}. 
    It shows that \textit{AdaPlan} consistently outperforms \textit{ReAct} on both in-domain and out-of-domain agent tasks across all models, demonstrating performance gains of 18.64\%, 13.58\%, 7.19\% on Qwen2.5-7B-Instruct, LLaMA3.1-8B-Instruct, and Qwen3-8B, respectively.}
\end{table*}

\begin{table*}[t] 
    \centering
	\resizebox{1\linewidth}{!}{
		\begin{tabular}{c|ccccccc}
            \toprule
                \multirow{2}{*}{Backbone Model} & \multirow{2}{*}{Architecture} & \multicolumn{1}{|c}{ALFWorld} & IQA & TextCraft & Wordle & \multicolumn{1}{|c}{BabyAI} & \multicolumn{1}{c}{MAZE}  \\
            \cline{3-8}
             & & \multicolumn{4}{|c}{\textit{In-Domain (ID)}} &  \multicolumn{2}{|c}{\textit{Out-of-Domain (OOD)}} \\
            \midrule
			\multirow{2}{*}{\makecell{Qwen2.5-7B\\-Instruct}} & Isolated & \multicolumn{1}{|c}{68.85} & 64.18 & 72.60 & 70.14 & \multicolumn{1}{|c}{58.29} & 52.07 \\
            & Unified & \multicolumn{1}{|c}{\textbf{70.80} ($\uparrow$ 2.83\%)} & \textbf{67.84} ($\uparrow$ 5.70\%) & \textbf{75.37} ($\uparrow$ 3.82\%) & \textbf{77.69} ($\uparrow$ 10.76\%) & \multicolumn{1}{|c}{\textbf{61.56} ($\uparrow$ 5.61\%)} & \textbf{57.93} ($\uparrow$ 11.25\%) \\
			\midrule
			\multirow{2}{*}{\makecell{LLaMA3.1-8B\\-Instruct}} & Isolated & \multicolumn{1}{|c}{71.87} & 70.83 & 60.96 & 71.05 & \multicolumn{1}{|c}{62.71} & 55.64 \\
            & Unified & \multicolumn{1}{|c}{\textbf{78.53} ($\uparrow$ 9.27\%)} & \textbf{72.78} ($\uparrow$ 2.75\%) & \textbf{64.76} ($\uparrow$ 6.23\%) & \textbf{79.61} ($\uparrow$ 12.05\%) & \multicolumn{1}{|c}{\textbf{68.24} ($\uparrow$ 8.82\%)} & \textbf{58.68} ($\uparrow$ 5.46\%) \\
            \midrule
            \multirow{2}{*}{Qwen3-8B} & Isolated & \multicolumn{1}{|c}{71.74} & 67.71 & 68.96 & 82.23 & \multicolumn{1}{|c}{60.55} & 51.49 \\
            & Unified & \multicolumn{1}{|c}{\textbf{72.51} ($\uparrow$ 1.07\%)} & \textbf{69.06} ($\uparrow$ 1.99\%) & \textbf{71.48} ($\uparrow$ 3.65\%) & \textbf{83.65} ($\uparrow$ 1.73\%) & \multicolumn{1}{|c}{\textbf{65.28} ($\uparrow$ 7.81\%)} & \textbf{56.62} ($\uparrow$ 9.96\%) \\
			\bottomrule
		\end{tabular}
	}
    \caption{\label{tab:coordination_ablation_detail}
    \textbf{Original scores for each benchmark of the planner-executor architecture analysis.} 
    It is the detailed version of \underline{\Cref{tab:coordination_ablation}}.
    The best scores of each model are in \textbf{bold}. 
    It shows that the \textit{unified architecture} consistently outperforms \textit{isolated architecture} on both in-domain and out-of-domain agent tasks across all models, with measured improvements of 6.48\%, 7.51\%, 3.96\% on Qwen2.5-7B-Instruct, LLaMA3.1-8B-Instruct, and Qwen3-8B.}
\end{table*}

\begin{table*}[th]
    \centering
    \resizebox{\textwidth}{!}{
    \begin{tabular}{cccc}
        \toprule
        \textbf{Dataset Category} & \textbf{Metric} & \textbf{ReAct} & \textbf{AdaPlan (Ours)} \\
        \midrule
        \multirow{2}{*}{\begin{tabular}[c]{@{}c@{}}Complex Interactive Tasks\\ (e.g., ALFWorld, BabyAI, TextCraft)\end{tabular}}
            & Avg. Token Consumption per Task & $\sim$560 & $\sim$450 \\
            & Inference Time (s) & $\sim$30 & $\sim$22 \\
        \midrule
        \multirow{2}{*}{\begin{tabular}[c]{@{}c@{}}Simple / Short Tasks\\ (e.g., Wordle, IQA, MAZE)\end{tabular}}
            & Avg. Token Consumption per Task & $\sim$180 & $\sim$300 \\
            & Inference Time (s) & $\sim$6s  & $\sim$8s \\
        \bottomrule
    \end{tabular}
    }
    \caption{\label{tab:cost_analysis}
    Cost analysis comparison between ReAct and AdaPlan.}  
\end{table*}

\subsection{Ablation Study Details}

\subsubsection{Original Performance Scores}

In this section, we report the original performance scores of the models on each benchmark during the training stage and training sequential order ablation, the agent paradigm analysis, as well as the planner-executor architecture analysis, as depicted in \Cref{tab:ablation_detail}, \Cref{tab:paradigm_ablation_detail} and \Cref{tab:coordination_ablation_detail}. 

\subsubsection{The Role of Each Stage}
\label{subsubsec:appendix_training_stage}

Here we provide the detailed observations of \Cref{subsubsec:training_stage} in the main content.

    \textbf{Removing Stage 1.} Stage 1 is designed to strengthen the models' ability to follow instructions when performing agent tasks. As shown in \Cref{tab:ablation} (\underline{2 $\rightarrow$ 3}), the removal of Stage 1 results in a performance drop of 4.20\% in overall model performance. This decline occurs because Stage 1 acts as the cornerstone for Stage 2. Without robust instruction-following behavior, the model struggles to adhere to the provided global plans, which are essential for delivering explicit guidance. As a result, the effectiveness of subsequent training stages is diminished to a certain extent.
    
    \textbf{Removing Stage 2.} Building upon Stage 1, Stage 2 focuses on optimizing the quality of generated global plans, thereby providing more effective high-level guidance for complex agent tasks. As indicated in \Cref{tab:ablation} (\underline{1 $\rightarrow$ 3}), eliminating Stage 2 results in a modest decline of 4.33\% in performance relative to the model trained with all three stages.
    
    \textbf{Removing Stage 3.} Stage 3 aims to optimize the coordination between the global planner and executor, thereby enhancing the model's end-to-end performance in agent tasks. As observed in \Cref{tab:ablation} (\underline{1 $\rightarrow$ 2}), excluding Stage 3 leads to a performance drop of 2.54\%. Nevertheless, owing to the presence of fully implemented Stage 1 and Stage 2, the performance gap relative to the model trained through all three stages remains narrow and relatively small.

\subsubsection{Declaration for \textit{\Cref{fig:line_chart}}} 
It is worth noting that when analyzing the evolution of reward scores for the global planner, the executor, and the end-to-end (E2E) performance using LLaMA3.1-8B-Instruct + PilotRL, we normalized all reward scores to the range $[0,1]$ for visualization and comparison purposes. The reward metrics include the following components:
\begin{itemize}
    \item \textbf{Global Planner}: This reward function (\Cref{eq:stage2_planning}) is introduced starting from Stage 2, and operates during Stage 2 (epoch 2 \& 3). In Stage 3, we only evaluate and record this metric without using it for model updates.
    \item \textbf{Executor}: This reward (\Cref{eq:stage1_adherence}) is used as the training objective solely in Stage 1. In the subsequent training stages, we continue to log its value for analysis, but it \textit{no longer} influences model updates.
    \item \textbf{End-to-End (E2E) Performance}: The reward based on end-to-end performance (\Cref{eq:stage1_e2e}) is evaluated throughout the entire training process and serves as a consistent metric for assessing overall system behavior.
\end{itemize}

\subsection{Analysis of Computational Overhead}
\label{appendix:computational_cost}

In this section, we further explore the tradeoffs of computational cost and task accuracy, as well as the training resource consumption.

\subsubsection{Inference Cost Analysis}

Here we provide the cost analysis of AdaPlan and ReAct in terms of average token consumption and inference time per task. 
While it is true that generating a global plan in AdaPlan introduces additional token consumption compared to the ReAct paradigm, this cost is justified by \textit{\textbf{(1)}} the substantial performance gains and \textit{\textbf{(2)}} the reduction in redundant execution steps. 
The generated formats of different agent paradigms are listed below:
\begin{itemize}
    \item \textbf{ReAct}: In the ReAct paradigm, the model primarily generates ``\texttt{Thought: ... Action: ...}'' per step. In fact, the token overhead of the ``\texttt{Thought}'' component is not negligible.
    \item \textbf{AdaPlan}: As for the AdaPlan paradigm, the model generates ``\texttt{Step 1: ... Step 2: ...}'', where each ``\texttt{Step}'' is clear and concise, with no excessive token consumption.
\end{itemize}

\begin{table*}[th]
\centering
\begin{tabular}{ccc}
\toprule
\textbf{Plan Update Frequency} & \textbf{Average Task Success (\%)} & \textbf{Average Total Steps} \\
\midrule
Every 5 Steps  & 53.28 ($\color{red} \downarrow$14.96) & 24.5 ($\color{red} \uparrow$8.7) \\
Every 2 Steps  & 62.57 ($\color{red} \downarrow$5.67) & 18.2 ($\color{red} \uparrow$2.4) \\
Every 1 Step (Ours) & 68.24 & 15.8 \\
\bottomrule
\end{tabular}
\caption{\label{tab:plan_update_frequency}
    Performance comparison of different plan update frequencies.} 
\end{table*}

\begin{table*}[htbp]
\centering
    \begin{tabular}{l|l}
    \toprule
    \textbf{Resource Item} & \textbf{Configuration / Value} \\
    \midrule
    Hardware & 32 $\times$ NVIDIA H20 GPUs (96GB memory each) \\
    Total training time & $\sim$18 hours \\
    \quad Stage 1: Executor Enhancement & $\sim$4 hours \\
    \quad Stage 2: Planner Cultivation & $\sim$8 hours \\
    \quad Stage 3: Joint Optimization & $\sim$6 hours \\
    Total token consumption & $\sim$4.6B tokens (4 epochs over 5,725 samples) \\
    GRPO optimization config & Rollout batch size = 64, using \texttt{vLLM} + \texttt{flash-attn-2}, \\
    \quad & achieving high generation throughput during sampling \\
    \bottomrule
    \end{tabular}
\caption{
\label{tab:training_cost}
Resource consumption for training LLaMA3.1-8B-Instruct}
\end{table*}

Compared to ReAct, AdaPlan incurs roughly a 20\%-30\% increase in tokens per step, and we provide the cost statistics in \Cref{tab:cost_analysis}, where we employed our trained Qwen2.5-7B-Instruct (Qwen2.5-7B-Instruct + PilotRL) for inference, and averaged the inference cost across benchmarks. The experiments are conducted on H20 GPUs, and we give a further analysis below:
\begin{itemize}[leftmargin=*]
    \item For \textit{complex interactive tasks}, AdaPlan reduces the total execution steps by 30\%-40\% (a typical case study is provided in \Cref{fig:case_study_ReAct} and \Cref{fig:case_study_PilotRL}, in which ReAct took 12+ steps with many wrong turns, while AdaPlan solved it in 6 steps), which offsets the cost of planning and even leads to a lower overall task completion cost (\texttt{Total Token Cost = Token per Step × Total Steps}). In these complex tasks, ReAct often suffers from ``trajectory redundancy'' or ``topic drift'', leading to longer execution paths or complete failure. Although AdaPlan spends more tokens per step, it reduces the total number of steps required to complete the task.
    \item For \textit{simple tasks}, the overhead is more pronounced. However, the absolute increase ($\sim$120 tokens) is negligible under the GPU inference, and the latency increase remains within acceptable ranges for most applications.
\end{itemize}

\subsubsection{Sensitivity Analysis on Plan Update Frequency}

We assessed the impact of plan update frequency using our trained LLaMA3.1-8B-Instruct (LLaMA3.1-8B-Instruct + PilotRL) on the BabyAI dataset. We compared three settings in terms of task success rate and average consumed steps:
\begin{itemize}[leftmargin=*]
    \item \textit{\underline{Low-Frequency}}: Plan updated every 5 steps.
    \item \textit{\underline{Medium-Frequency}}: Plan updated every 2 steps.
    \item \textit{\underline{High-Frequency}}: Plan updated every 1 step, which is the strategy applied in our method.
\end{itemize}
The results are shown in \Cref{tab:plan_update_frequency}, and we give a further clarification below:
\begin{itemize}[leftmargin=*]
    \item Higher frequency leads to \textit{\textbf{higher success rates}} due to the faster error correction. The 14.96\% drop in success rate when moving from ``Every 1 Step'' to ``Every 5 Steps'' indicates that relying on outdated plans (stale context) leads to suboptimal decisions and catastrophic failures in complex environments. Frequent updates allow the agent to correct errors before they compound.
    \item The high-frequency approach (ours) actually results in \textit{\textbf{the most efficient execution path}}, which is shorter than the ``Every 2 Steps'' strategy (18.2 steps) and the ``Every 5 Steps'' strategy (24.5 steps). This implies that real-time replanning helps the model avoid dead-ends, backtrack less, and navigate the solution space more directly. Although the model spends slightly more compute time thinking per step, it saves a substantial amount of time in execution.
\end{itemize}

\subsubsection{Training Resource Consumption}

Here we discuss the total token consumption and actual training duration. For clarity, we report the actual resource consumption for training LLaMA3.1-8B-Instruct, as described in \Cref{tab:training_cost} (and similar for Qwen2.5-7B-Instruct and Qwen3-8B, with parameter scales of $\sim$7-8B).

\subsection{Further Analysis on \textit{Frontier Models}}

When conducting our main experiment, we employ DeepSeek-V3 as the frontier model for three roles:
\begin{enumerate}
    \item \underline{\textit{Environment Simulator:}} As described in \Cref{subsec:experimental_setup}, the frontier model simulates real-world environmental behaviors for its reliability and computational efficiency, where we employ the approach from ~\citet{sun2025zerosearch}.
    \item \underline{\textit{Global Plan Generation:}} As stated in \Cref{subsubsec:training_stage1}, the frontier model generates the initial global plans in Stage 1 during the entire training process of PilotRL.
    \item \underline{\textit{Evaluation:}} The frontier model is adopted as the judge in the \textit{LLM-as-Judge} paradigm to verify key metrics (e.g., adherence degree, global plan quality) during the RL process, and evaluate the model's E2E performance.
\end{enumerate}

In this section, we present an in-depth analysis of the use of frontier models in our experimental pipeline. This includes a systematic enumeration of their advantages, ablation studies across different frontier models, and human expert evaluation of the judgment results generated by these models.

\subsubsection{Advantages of Frontier Models}

We employ frontier models in our \ours pipeline for the following reasons:

\begin{table*}[tb]
    \centering
    \resizebox{\textwidth}{!}{
        \begin{tabular}{c|c|cccccccc}
            \toprule
                \multirow{2}{*}{Frontier Model} & \multirow{2}{*}{Method} & \multirow{2}{*}{w/o Plan.} & \multicolumn{1}{|c}{ALFWorld} & IQA & TextCraft & Wordle & \multicolumn{1}{|c}{BabyAI} & \multicolumn{1}{c|}{MAZE} & \multirow{2}{*}{\textbf{Avg.}} \\
            \cline{4-9}
             & & & \multicolumn{4}{|c}{\textit{In-Domain (ID)}} &  \multicolumn{2}{|c|}{\textit{Out-of-Domain (OOD)}} & \\
            \midrule
            \multirow{6}{*}{LLaMA3.1-70B-Instruct} & Naive Response & \ding{55} & \multicolumn{1}{|c}{45.92} & 32.18 & 27.61 & 31.45 & \multicolumn{1}{|c}{37.63} & 30.54 & \multicolumn{1}{|c}{34.22} \\
             & ReAct & \ding{55} & \multicolumn{1}{|c}{49.33} & 34.82 & 31.24 & 37.68 & \multicolumn{1}{|c}{40.87} & 34.71 & \multicolumn{1}{|c}{38.11} \\
             & + MPO & \ding{52} & \multicolumn{1}{|c}{64.57} & 55.39 & 49.52 & 53.41 & \multicolumn{1}{|c}{51.12} & 46.43 & \multicolumn{1}{|c}{53.41} \\
             & SFT & \ding{52} & \multicolumn{1}{|c}{\underline{64.78}} & 60.12 & \underline{70.34} & \underline{71.89} & \multicolumn{1}{|c}{52.45} & 44.15 & \multicolumn{1}{|c}{\underline{60.62}} \\
             & Vanilla RL & \ding{55} & \multicolumn{1}{|c}{62.73} & \underline{61.54} & 67.98 & 68.02 & \multicolumn{1}{|c}{\underline{55.89}} & \underline{47.31} & \multicolumn{1}{|c}{60.58} \\
             & \textbf{\ours (ours)} & \ding{52} & \multicolumn{1}{|c}{\textbf{68.05}} & \textbf{64.61} & \textbf{72.63} & \textbf{74.42} & \multicolumn{1}{|c}{\textbf{58.79}} & \textbf{54.67} & \multicolumn{1}{|c}{\textbf{65.53}} \\
             \midrule
             \multirow{6}{*}{GPT-4o} & Naive Response & \ding{55} & \multicolumn{1}{|c}{51.94} & 36.15 & 33.62 & 35.98 & \multicolumn{1}{|c}{43.62} & 34.57 & \multicolumn{1}{|c}{39.31} \\
             & ReAct & \ding{55} & \multicolumn{1}{|c}{53.38} & 38.32 & 35.71 & 41.19 & \multicolumn{1}{|c}{45.31} & 38.26 & \multicolumn{1}{|c}{42.03} \\
             & + MPO & \ding{52} & \multicolumn{1}{|c}{68.56} & 59.87 & 53.05 & 58.03 & \multicolumn{1}{|c}{54.63} & 50.92 & \multicolumn{1}{|c}{57.51} \\
             & SFT & \ding{52} & \multicolumn{1}{|c}{\underline{68.79}} & 64.12 & \underline{74.36} & \underline{75.88} & \multicolumn{1}{|c}{56.43} & 48.17 & \multicolumn{1}{|c}{64.63} \\
             & Vanilla RL & \ding{55} & \multicolumn{1}{|c}{66.72} & \underline{65.53} & 71.94 & 72.51 & \multicolumn{1}{|c}{\underline{59.87}} & \underline{51.34} & \multicolumn{1}{|c}{\underline{64.65}} \\
             & \textbf{\ours (ours)} & \ding{52} & \multicolumn{1}{|c}{\textbf{72.03}} & \textbf{68.62} & \textbf{76.61} & \textbf{78.45} & \multicolumn{1}{|c}{\textbf{62.81}} & \textbf{58.76} & \multicolumn{1}{|c}{\textbf{69.55}} \\
            \bottomrule
        \end{tabular}
    }
    \caption{\label{tab:llm_judge_ablation} 
    \textbf{Ablation for the alternative of frontier models.} 
    Here we utilize \textbf{\textit{Qwen2.5-7B-Instruct}} as the same backbone model. 
    ``w/o Plan.'' indicates whether the inference paradigm includes global planning as a mechanism for providing explicit guidance. 
    The best and second best of each model are in \textbf{bold} and \underline{underlined}.}
\end{table*}

\begin{table*}[t] 
    \centering
	\resizebox{1\linewidth}{!}{
		\begin{tabular}{c|cccccc|c}
            \toprule
                \multirow{2}{*}{Backbone Model} & \multicolumn{1}{|c}{ALFWorld} & IQA & TextCraft & Wordle & \multicolumn{1}{|c}{BabyAI} & \multicolumn{1}{c|}{MAZE} & \multirow{2}{*}{\textbf{Avg.}} \\
            \cline{2-7}
             & \multicolumn{4}{|c}{\textit{In-Domain (ID)}} &  \multicolumn{2}{|c|}{\textit{Out-of-Domain (OOD)}} & \\
            \midrule
			Qwen2.5-7B-Instruct & \multicolumn{1}{|c}{1.00 (30/30)} & 0.97 (29/30) & 0.93 (28/30) & 1.00 (30/30) & \multicolumn{1}{|c}{1.00 (30/30)} & 1.00 (30/30) & 0.98 \\
			\midrule
			LLaMA3.1-8B-Instruct & \multicolumn{1}{|c}{0.97 (29/30)} & 0.93 (28/30) & 0.97 (29/30) & 1.00 (30/30) & \multicolumn{1}{|c}{1.00 (30/30)} & 1.00 (30/30) & 0.98 \\
            \midrule
            Qwen3-8B & \multicolumn{1}{|c}{1.00 (30/30)} & 1.00 (30/30) & 0.90 (27/30) & 1.00 (30/30) & \multicolumn{1}{|c}{0.97 (29/30)} & 1.00 (30/30) & 0.98 \\
			\bottomrule
		\end{tabular}
	}
    \caption{\label{tab:meta-evaluation}
    \textbf{Meta-evaluation results of the frontier model \textit{DeepSeek-V3}'s judge.} We sample 30 instances per dataset, and observe a high evaluation accuracy across all the benchmarks.}
\end{table*}

\textbf{\textit{Reliability.}} 
The frontier models have acquired extensive commonsense knowledge and reasoning capabilities during training through exposure to trillions of high-quality tokens. 
\begin{itemize}
    \item For \textit{environmental simulation}, it has strong semantic task understanding, enabling it to directly infer both the agent’s current state (e.g., the spatial configuration, locations of objects), and the desired goal state from the multi-turn dialogue context in benchmarks like ALFWorld and BabyAI. In contrast, native simulators, while offering high-fidelity interactive environments, provide only low-level perceptual feedback (e.g., ``room layout'', ``object positions'') without explicit semantic interpretation of the task objective. 
    \item For \textit{global plan generation}, the frontier model can provide a stable, high-quality prior that guides the agent toward semantically coherent planning behavior during the initial training phase, which serves as a crucial scaffold for subsequent training stages. 
    \item For \textit{evaluation}, LLM-as-Judge enables semantic equivalence judgment and logical coherence assessment, aligning better with the open-ended nature of agent tasks. It is now widely adopted in major agent benchmarks where rule-based metrics fail to capture semantic correctness~\cite{zheng2023judging,li2025generation}. 
    Moreover, when the frontier model evaluates task completion, we additionally provide reference trajectories sourced from \citet{song2024agentbank} and \citet{xi2024agentgym}, which further supply a concrete reference standard to guide and calibrate its judgments, including task completion and solution efficiency.
\end{itemize}

\textbf{\textit{Computational Efficiency.}} 
When \textit{simulating real-world environmental behaviors}, the frontier model offers significantly lower computational overhead during inference compared to executing interactions in a physical simulator. Native simulators like ALFWorld rely on graphics engines such as Unity~\cite{nicoll2019unity} or AI2-THOR~\cite{kolve2017ai2}, which necessitate loading 3D scenes, rendering visual inputs, and maintaining complex state machines. This would heavily increase the computational overhead during the rollout phase (simulating multi-turn interactions with the environment) in RL training. By contrast, the frontier model operates purely on textual representations and can be deployed efficiently on standard hardware, enabling large-scale experimentation even under constrained computational budgets. 
As for \textit{evaluation}, LLM-as-Judge has become a standard evaluation approach for LLM agents, as human evaluation is costly and unscalable for long-horizon, open-ended tasks, where the task solutions can take flexible forms. 

\subsubsection{Ablations of the Frontier Models}

Here we use Qwen2.5-7B-Instruct as the backbone model and conduct experiments by replacing \textbf{DeepSeek-V3} (utilized in main experiments) with open-sourced alternative \textbf{LLaMA3.1-70B-Instruct} and close-soured alternative \textbf{GPT-4o}. 

The experimental results are shown in \Cref{tab:llm_judge_ablation}, and it can be observed that, due to differences in model scoring preferences, there is indeed some variation in scores under the LLM-as-Judge paradigm. Nevertheless, our PilotRL consistently outperforms the other baselines overall, which shows the robustness of our design.

\subsubsection{Meta-Evaluation of Frontier Models}

We also employ human evaluation to ensure the judge from the frontier model has a high agreement with expert judge. 
Specifically, we sample 30 instances per dataset, collect the generation results from Qwen2.5-7B-Instruct, LLaMA3.1-8B-Instruct and Qwen3-8B, and then report the judgment correctness of the frontier model DeepSeek-V3 (employed in our main experiment).

As depicted in \Cref{tab:meta-evaluation}, our human meta-evaluation study demonstrates that the frontier model evaluator achieves approximately 98\% agreement with human experts. 
The model’s judgment slightly deviates from human experts for the TextCraft problems since the task is inherently creative text generation, and its evaluation relies on subjective aesthetics to some extent. 
However, the LLM-as-Judge paradigm still achieves 90\% or higher accuracy across all datasets.

\subsection{Implementation Details}

To ensure reproducibility, we have provided detailed descriptions on the environment, hardware configurations, and hyperparameters.

\subsubsection{Environment}
We conduct experiments by utilizing the following core libraries and their respective versions: torch=2.5.1, CUDA\_version=12.4, ray=2.40.0, vllm=0.7.3, verl=0.2.0.post2, transfomrers=4.49.0, datasets=3.3.2, tqdm=4.40.0, flash-attn=2.5.8, pyarrow=19.0.1, tensordict=0.5.0. 

\subsubsection{Hardware Configurations}
Experiments are conducted using 32 NVIDIA H20 GPUs with 96GB memory.

\begin{table}[th]
	\centering
	\resizebox{1\linewidth}{!}{
	\begin{tabular}{l|cc}
		\toprule
		\textbf{Hyperparameters} & \textbf{RL} & \textbf{SFT}  \\
		\midrule
		Learning rate & 1e\mbox{-}6 & 2e\mbox{-}5 \\
		LR scheduler & Cosine & Cosine \\
        Warm-up ratio & 0.03 & 0.03 \\
		Weight decay & 0.0 & 0.0 \\
		Gradient clipping & 1.0 & 1.0 \\
		Optimizer  & \multicolumn{2}{c}{AdamW ($\beta_1=0.9, \beta_2=0.999$)}\\
		Training batch size & 256 & 256 \\
        Rollout batch size & 64 & -- \\
        Rollout engine & \texttt{vLLM} & -- \\
		Mixed precision training & BF16 & BF16 \\
		\bottomrule
	\end{tabular}
    }
    \caption{\label{tab:hyper}
    Hyperparameters for RL and SFT training.}
\end{table}

\subsubsection{Hyperparameters Details}

The detailed hyperparameters of reinforcement learning (RL) and supervised fine-tuning (SFT) training are summarized in \Cref{tab:hyper}, where our \ours and the \textit{Vanilla RL} baseline  are trained under the RL setting.





\section{Prompts}
\label{appendix:prompts}

In this section we present the prompts used throughout our pipeline in \ours. 
Only the English version is presented due to LaTeX compilation issues with non-English languages.

\begin{tcolorbox}[breakable, title=\textbf{Prompt: Global Plan Generation - ALFWorld}, colback=gray!7, colframe=gray!50!black, boxrule=1pt]
Based on the task description, the previous global plan, and accumulated observation of agent interactions with the environment, please generate all possible step-by-step global plans, which serve as high-level, natural guidance to assist in planning.
Maintain the plan for all steps preceding the execution step index, while selectively modifying the plan for steps following the execution step index. 

\vspace{0.5cm}

For house holding task, the action list you can take:
\begin{enumerate}
    \item go to recep
    \item task obj from recep
    \item put obj in/on recep
    \item open recep
    \item close recep
    \item toggle obj recep
    \item clean obj with recep
    \item heat obj with recep
    \item cool obj with rece
\end{enumerate}
where obj and recep correspond to objects and receptacles.

\vspace{0.5cm}

\textbf{\# Task}\\
\{task\}

\vspace{0.5cm}

\textbf{\# Previous Global Plan}\\
\{global\_plan\} [optional]

\vspace{0.5cm}

\textbf{\# Execution Step Index}\\
\{execution\_step\_index\}

\vspace{0.5cm}

\textbf{\# Accumulated Observation}\\
\{observation\} [optional]

\vspace{0.5cm}

\hdashrule{\textwidth}{1pt}{3pt}

\vspace{0.5cm}

\textbf{Output Format:}  
\begin{verbatim}
```json
["
Step 1: ...
Step 2: ...
...
", ...]
'''
\end{verbatim}

\end{tcolorbox}

\begin{tcolorbox}[breakable, title=\textbf{Prompt: Global Plan Generation - IQA}, colback=gray!7, colframe=gray!50!black, boxrule=1pt]
Based on the task description, the previous global plan, and accumulated observation of agent interactions with the environment, please generate all possible step-by-step global plans, which serve as high-level, natural guidance to assist in planning.
Maintain the plan for all steps preceding the execution step index, while selectively modifying the plan for steps following the execution step index. 

\vspace{0.5cm}

For interactive QA task, the action list you can take:
\begin{enumerate}
    \item move ahead
    \item turn left
    \item turn right
    \item open obj
    \item answer [True]/[False]
\end{enumerate}
where obj correspond to objects. 

\vspace{0.5cm}

\textbf{\# Task}\\
\{task\}

\vspace{0.5cm}

\textbf{\# Previous Global Plan}\\
\{global\_plan\} [optional]

\vspace{0.5cm}

\textbf{\# Execution Step Index}\\
\{execution\_step\_index\}

\vspace{0.5cm}

\textbf{\# Accumulated Observation}\\
\{observation\} [optional]

\vspace{0.5cm}

\hdashrule{\textwidth}{1pt}{3pt}

\vspace{0.5cm}

\textbf{Output Format:}  
\begin{verbatim}
```json
["
Step 1: ...
Step 2: ...
...
", ...]
'''
\end{verbatim}

\end{tcolorbox}

\begin{tcolorbox}[breakable, title=\textbf{Prompt: Global Plan Generation - TextCraft}, colback=gray!7, colframe=gray!50!black, boxrule=1pt]
You are given a few useful crafting recipes to craft items in Minecraft. Craft command can be understood as follows: craft [target] using [ingredients], where target is item/object generated by the craft command as output and ingredient are the inputs. You are given an agent that can ``craft'' or ``fetch'' objects. You can take the help of crafting commands below to create new objects.
Based on the task description, the previous global plan, and accumulated observation of agent interactions with the environment, please generate all possible step-by-step global plans, which serve as high-level, natural guidance to assist in planning. 
Maintain the plan for all steps preceding the execution step index, while selectively modifying the plan for steps following the execution step index. 
Each global plan can use at most ONE of the provided crafting commands.

\vspace{0.5cm}

\textbf{\# Task}\\
\{task\}

\vspace{0.5cm}

\textbf{\# Previous Global Plan}\\
\{global\_plan\} [optional]

\vspace{0.5cm}

\textbf{\# Execution Step Index}\\
\{execution\_step\_index\}

\vspace{0.5cm}

\textbf{\# Accumulated Observation}\\
\{observation\} [optional]

\vspace{0.5cm}

\hdashrule{\textwidth}{1pt}{3pt}

\vspace{0.5cm}

\textbf{Output Format:}  
\begin{verbatim}
```json
["
Step 1: ...
Step 2: ...
...
", ...]
'''
\end{verbatim}

\end{tcolorbox}

\begin{tcolorbox}[breakable, title=\textbf{Prompt: Global Plan Generation - Wordle}, colback=gray!7, colframe=gray!50!black, boxrule=1pt]
You are an expert wordle player. 
Based on the task description, the previous global plan, and accumulated observation of agent interactions with the environment, please generate all possible step-by-step global plans for the wordle task, which serve as high-level, natural guidance to assist in planning. 
Maintain the plan for all steps preceding the execution step index, while selectively modifying the plan for steps following the execution step index. 
Your objective is to guess a hidden 5 letter word. You have 6 attempts to guess it correctly and you should try to guess it in as few attempts as possible. When guessing the word, you should format your word as a space separated sequence of letters, like ``s h i r e'' for example. After guessing the word, you will receive feedback from the game environment in the form of a sequence of 5 space separated letters like ``b y g g b'', where each letter indicates some information about the hidden word. The environment will return one of three letters - ``b'', ``g'', or ``y'' – for each letter in the word you guessed. Here is the meaning of each letter: 

\begin{itemize}
    \item ``b'': If the environment returns a ``b'', it means that the letter at that position in your guessed word is not in the hidden word.
    \item ``y'': If the environment returns a ``y'', it means that the letter at that position in your guessed word is in the hidden word but is not in the correct position.
    \item ``g'': If the environment returns a ``g'', it means that the letter at that position in your guessed word is in the hidden word and is in the correct position.
\end{itemize}

\vspace{0.5cm}

\textbf{\# Task}\\
\{task\}

\vspace{0.5cm}

\textbf{\# Previous Global Plan}\\
\{global\_plan\} [optional]

\vspace{0.5cm}

\textbf{\# Execution Step Index}\\
\{execution\_step\_index\}

\vspace{0.5cm}

\textbf{\# Accumulated Observation}\\
\{observation\} [optional]

\vspace{0.5cm}

\hdashrule{\textwidth}{1pt}{3pt}

\vspace{0.5cm}

\textbf{Output Format:}  
\begin{verbatim}
```json
["
Step 1: ...
Step 2: ...
...
", ...]
'''
\end{verbatim}

\end{tcolorbox}

\begin{tcolorbox}[breakable, title=\textbf{Prompt: Global Plan Generation - BabyAI}, colback=gray!7, colframe=gray!50!black, boxrule=1pt]
You are an exploration master that wants to finish every goal you are given. 
You are placed in a room and you need to accomplish the given goal with actions. 
Based on the task description, the previous global plan, and accumulated observation of agent interactions with the environment, please generate all possible step-by-step global plans, which serve as high-level, natural guidance to assist in planning. Maintain the plan for all steps preceding the execution step index, while selectively modifying the plan for steps following the execution step index. 

\vspace{0.5cm}

The action list you can take:
\begin{enumerate}
    \item turn right
    \item turn left
    \item move forward
    \item go to $<$obj$>$ $<$id$>$
    \item pick up $<$obj$>$ $<$id$>$
    \item go through $<$door$>$ $<$id$>$: $<$door$>$ must be an open door. 
    \item toggle and go through $<$door$>$ $<$id$>$: $<$door$>$ can be a closed door or a locked door. If you want to open a locked door, you need to carry a key that is of the same color as the locked door.
    \item toggle: there is a closed or locked door right in front of you and you can toggle it.
\end{enumerate}
where $<$obj$>$ and $<$id$>$ correspond to objects and index number. 

\vspace{0.5cm}

\textbf{\# Task}\\
\{task\}

\vspace{0.5cm}

\textbf{\# Previous Global Plan}\\
\{global\_plan\} [optional]

\vspace{0.5cm}

\textbf{\# Execution Step Index}\\
\{execution\_step\_index\}

\vspace{0.5cm}

\textbf{\# Accumulated Observation}\\
\{observation\} [optional]

\vspace{0.5cm}

\hdashrule{\textwidth}{1pt}{3pt}

\vspace{0.5cm}

\textbf{Output Format:}  
\begin{verbatim}
```json
["
Step 1: ...
Step 2: ...
...
", ...]
'''
\end{verbatim}

\end{tcolorbox}

\begin{tcolorbox}[breakable, title=\textbf{Prompt: Global Plan Generation - MAZE}, colback=gray!7, colframe=gray!50!black, boxrule=1pt]
You are an expert maze solver. 
Your objective is to reach the goal in as few steps as possible. 
Based on the task description, the previous global plan, and accumulated observation of agent interactions with the environment, please generate all possible step-by-step global plans, which serve as high-level, natural guidance to assist in planning. Maintain the plan for all steps preceding the execution step index, while selectively modifying the plan for steps following the execution step index. 
Your objective is to reach the goal in as few steps as possible. When you move right, you increase your \texttt{y} position by 1. When you move down, you increase your \texttt{x} position by 1.

\vspace{0.5cm}

The action list you can take:

\begin{enumerate}
    \item move up
    \item move down
    \item move left
    \item move right
\end{enumerate}

\vspace{0.5cm}

\hdashrule{\textwidth}{1pt}{3pt}

\vspace{0.5cm}

For instance, given the current environment state:
The goal is at position 8, 6. Your current position is at position 1, 1. There are walls to your left, above you, below you. The index of already executed steps is \texttt{0}. The possible global plans could be: 

[`` \\
Step 1: move right (from 1, 1 to 1, 2) \\
Step 2: move right (from 1, 2 to 1, 3) \\
Step 3: move right (from 1, 3 to 1, 4) \\
Step 4: move down (from 1, 4 to 2, 4) \\
Step 5: move down (from 2, 4 to 3, 4) \\
Step 6: move down (from 3, 4 to 4, 4) \\
Step 7: move down (from 4, 4 to 5, 4) \\
Step 8: move down (from 5, 4 to 6, 4) \\
Step 9: move down (from 6, 4 to 7, 4) \\
Step 10: move down (from 7, 4 to 8, 4) \\
Step 11: move right (from 8, 4 to 8, 5) \\
Step 12: move right (from 8, 5 to 8, 6) \\
'', ...]

\vspace{0.5cm}

\hdashrule{\textwidth}{1pt}{3pt}

\vspace{0.5cm}

\textbf{\# Task}\\
\{task\}

\vspace{0.5cm}

\textbf{\# Previous Global Plan}\\
\{global\_plan\} [optional]

\vspace{0.5cm}

\textbf{\# Execution Step Index}\\
\{execution\_step\_index\}

\vspace{0.5cm}

\textbf{\# Accumulated Observation}\\
\{observation\} [optional]

\vspace{0.5cm}

\hdashrule{\textwidth}{1pt}{3pt}

\vspace{0.5cm}

\textbf{Output Format:}  
\begin{verbatim}
```json
["
Step 1: ...
Step 2: ...
...
", ...]
'''
\end{verbatim}

\end{tcolorbox}

\begin{tcolorbox}[breakable, title=\textbf{Prompt: Global Plan Selection (for the \textit{generate-then-select} strategy)}, colback=gray!7, colframe=gray!50!black, boxrule=1pt]
You are given several global plans serving as high-level, natural guidance to assist in planning. 
Based on the task description, accumulated observation of agent interactions with the environment, and the current index of execution step, 
please select the most suitable global plan from all available global plans for task completion. \\

When you select the global plan, consider evaluating the following aspects to identify the optimal choice based on comprehensive criteria:

\begin{enumerate}
    \item \textbf{\textit{\underline{Correctness}}}: Does the global plan correctly and accurately address the task requirements?
    \item \textbf{\textit{\underline{Executability}}}: Is the global plan clearly structured, easy to interpret, and are the individual steps logically sound and actionable?
    \item \textbf{\textit{\underline{Standardization}}}: Does the global plan adhere to a consistent and standardized format?
\end{enumerate}

\vspace{0.5cm}

\textbf{\# Task}\\
\{task\}

\vspace{0.5cm}

\textbf{\# Available Global Plans}\\
\{global\_plans\}

\vspace{0.5cm}

\textbf{\# Execution Step Index}\\
\{execution\_step\_index\}

\vspace{0.5cm}

\textbf{\# Accumulated Observation}\\
\{observation\} [optional]

\end{tcolorbox}

\begin{tcolorbox}[breakable, title=\textbf{Prompt: Global Plan Quality Evaluation (for \Cref{eq:stage2_planning})}, colback=gray!7, colframe=gray!50!black, boxrule=1pt]
Please act as a professional guidance evaluator and judge the given global plan across the following three dimensions: 

\begin{enumerate}
    \item \textbf{\textit{\underline{Correctness}}}: Based on the environment's feedback on the agent's actions in response to the current global plan guidance, does the global plan accurately fulfill the task requirements?
    \item \textbf{\textit{\underline{Executability}}}: Based on the agent's adherence to the global plan, is the global plan clear, easy to understand, and are the steps reasonable?
    \item \textbf{\textit{\underline{Standardization}}}: Does the global plan adhere to a consistent and standardized format?
\end{enumerate}

For each dimension, please score the global plan on a scale of 1 to 5, where 1 indicates poor performance and 5 indicates excellent performance, and explain the reason.

\vspace{0.5cm}

\textbf{\# Task}\\
\{task\}

\vspace{0.5cm}

\textbf{\# Global Plan}\\
\{global\_plan\}

\vspace{0.5cm}

\textbf{\# Execution Step Index}\\
\{execution\_step\_index\}

\vspace{0.5cm}

\textbf{\# Accumulated Observation}\\
\{observation\} [optional]

\vspace{0.5cm}

\hdashrule{\textwidth}{1pt}{3pt}

\vspace{0.5cm}

\textbf{Output Format:}  
\begin{verbatim}
```json
{
    "correctness_score": xxx,
    "correctness_reason": "...",
    "executability_score": xxx,
    "executability_reason": "...",
    "standardization_score": xxx,
    "standardization_reason": "..."
}
'''
\end{verbatim}

\end{tcolorbox}

\begin{tcolorbox}[breakable, title=\textbf{Prompt: Environmental Feedback}, colback=gray!7, colframe=gray!50!black, boxrule=1pt]
Based on the task description and the reference agent-environment interaction in which the agent has finally accomplished the task, please generate the environmental feedback for the agent's action and determine whether the current action has reached the final goal. If the agent's action has reached the final goal, please output ``Task Completed!''; 
else, the feedback should be in the following format:
``Observation: ...''

\vspace{0.5cm}

\textbf{\# Task}\\
\{task\}

\vspace{0.5cm}

\textbf{\# Reference Interaction}\\
\{ref\_interaction\}

\vspace{0.5cm}

\textbf{\# Previous Observation}\\
\{observation\} [optional]

\vspace{0.5cm}

\textbf{\# Agent Action}\\
\{agent\_action\}

\end{tcolorbox}

\begin{tcolorbox}[breakable, title=\textbf{Prompt: Execution Generation - ALFWorld}, colback=gray!7, colframe=gray!50!black, boxrule=1pt]
Interact with a household to solve a task. Imagine you are an intelligent agent in a household environment and your target is to perform actions to complete the task goal. At the beginning of your interactions, you will be given the detailed description of the current environment and your goal to accomplish.
For each of your turn, you will be given the observation of the last turn. You should choose from two actions: ``Thought'' or ``Action''. If you choose ``Thought'', you should first think about the current condition and plan for your future actions, and then output your action in this turn. Your output must strictly follow this format: ``Thought: your thoughts. Action: your next action''; If you choose ``Action'', you should directly output the action in this turn. Your output must strictly follow this format: ``Action: your next action''.

\vspace{0.5cm}

For house holding task, the action list you can take:
\begin{enumerate}
    \item go to recep
    \item task obj from recep
    \item put obj in/on recep
    \item open recep
    \item close recep
    \item toggle obj recep
    \item clean obj with recep
    \item heat obj with recep
    \item cool obj with rece
\end{enumerate}
where obj and recep correspond to objects and receptacles.

\vspace{0.5cm}

Reminder:
\begin{enumerate}
    \item The action is restricted to those listed as available. Actions not included in the provided list are considered invalid.
    \item Think when necessary, but prioritize direct action wherever possible throughout the process.
\end{enumerate}

\vspace{0.5cm}

\textbf{\# Example}\\
\{example\}

\vspace{0.5cm}

\textbf{\# Task}\\
\{task\}

\vspace{0.5cm}

\textbf{\# Global Plan}\\
\{global\_plan\}

\vspace{0.5cm}

\textbf{\# Previous Observation}\\
\{observation\} [optional]

\end{tcolorbox}

\begin{tcolorbox}[breakable, title=\textbf{Prompt: Execution Generation - IQA}, colback=gray!7, colframe=gray!50!black, boxrule=1pt]
Imagine you are an intelligent agent in a dynamic visual environment and your target is to perform actions to complete the task goal. At the beginning of your interactions, you will be given the detailed description of the current environment and your goal to accomplish. 
For each of your turn, you will be given the observation of the last turn. You should choose from two actions: ``Thought'' or ``Action''. If you choose ``Thought'', you should first think about the current condition and plan for your future actions, and then output your action in this turn. Your output must strictly follow this format: ``Thought: your thoughts. Action: your next action''; If you choose ``Action'', you should directly output the action in this turn. Your output must strictly follow this format: ``Action: your next action''.

\vspace{0.5cm}

The action list you can take:
\begin{enumerate}
    \item move ahead
    \item turn left
    \item turn right
    \item open obj
    \item answer [True]/[False]
\end{enumerate}
where obj correspond to objects. 

\vspace{0.5cm}

Reminder:
\begin{enumerate}
    \item The action is restricted to those listed as available. Actions not included in the provided list are considered invalid.
    \item Think when necessary, but prioritize direct action wherever possible throughout the process.
\end{enumerate}

\vspace{0.5cm}

\textbf{\# Example}\\
\{example\}

\vspace{0.5cm}

\textbf{\# Task}\\
\{task\}

\vspace{0.5cm}

\textbf{\# Global Plan}\\
\{global\_plan\}

\vspace{0.5cm}

\textbf{\# Previous Observation}\\
\{observation\} [optional]

\end{tcolorbox}

\begin{tcolorbox}[breakable, title=\textbf{Prompt: Execution Generation - TextCraft}, colback=gray!7, colframe=gray!50!black, boxrule=1pt]
You are given a few useful crafting recipes to craft items in Minecraft. Crafting commands are of the format ``craft [target object] using [input ingredients]''. Every round I will give you an observation, you have to respond to an action based on the state and instruction. 
You should choose from two actions: ``Thought'' or ``Action''. If you choose ``Thought'', you should first think about the current condition and plan for your future actions, and then output your action in this turn. Your output must strictly follow this format: ``Thought: your thoughts. Action: your next action''; If you choose ``Action'', you should directly output the action in this turn. Your output must strictly follow this format: ``Action: your next action''. 
For your ``Action'', you can ``get'' an object (ingredients) from the inventory or the environment, look up the game ``inventory'' by inventory, or ``craft'' (target) using any of the crafting commands. You can use ONLY these crafting commands provided, do not use your own crafting commands. However, if the crafting command uses a generic ingredient like ``planks'', you can use special types of the same ingredient e.g. dark oak ``planks'' in the command instead. For any other natural language or thoughts, use prefix 'Thought:'.

\vspace{0.5cm}

Reminder:
\begin{enumerate}
    \item The action is restricted to those listed as available. Actions not included in the provided list are considered invalid.
    \item Think when necessary, but prioritize direct action wherever possible throughout the process.
\end{enumerate}

\vspace{0.5cm}

\textbf{\# Example}\\
\{example\}

\vspace{0.5cm}

\textbf{\# Crafting Commands and Goal}\\
\{task\}

\vspace{0.5cm}

\textbf{\# Global Plan}\\
\{global\_plan\}

\vspace{0.5cm}

\textbf{\# Previous Observation}\\
\{observation\} [optional]

\end{tcolorbox}

\begin{tcolorbox}[breakable, title=\textbf{Prompt: Execution Generation - Wordle}, colback=gray!7, colframe=gray!50!black, boxrule=1pt]
You are an expert wordle player. Welcome to the game of Wordle. Your objective is to guess a hidden 5 letter word. You have 6 attempts to guess it correctly and you should try to guess it in as few attempts as possible. When guessing the word, you should format your word as a space separated sequence of letters, like ``s h i r e'' for example. After guessing the word, you will receive feedback from the game environment in the form of a sequence of 5 space separated letters like ``b y g g b'', where each letter indicates some information about the hidden word. The environment will return one of three letters - ``b'', ``g'', or ``y'' – for each letter in the word you guessed. Here is the meaning of each letter: 

\begin{itemize}
    \item ``b'': If the environment returns a ``b'', it means that the letter at that position in your guessed word is not in the hidden word.
    \item ``y'': If the environment returns a ``y'', it means that the letter at that position in your guessed word is in the hidden word but is not in the correct position.
    \item ``g'': If the environment returns a ``g'', it means that the letter at that position in your guessed word is in the hidden word and is in the correct position.
\end{itemize}

For each of your turn, you will be given the observation of the last turn. You should choose from two actions: ``Thought'' or ``Action''. If you choose ``Thought'', you should first think about the current condition and plan for your future actions, and then output your action in this turn. Your output must strictly follow this format: ``Thought: your thoughts. Action: your next action''; If you choose ``Action'', you should directly output the action in this turn. Your output must strictly follow this format: ``Action: your next action''.

\vspace{0.5cm}

Reminder:
\begin{enumerate}
    \item The output format of the action should be a sequence of 5 individual letters, each separated by a space, such as ``s h i r e''. 
    Any other formats are considered invalid.
    \item Think when necessary, but prioritize direct action wherever possible throughout the process.
\end{enumerate}

\vspace{0.5cm}

\textbf{\# Example}\\
\{example\}

\vspace{0.5cm}

\textbf{\# Task}\\
\{task\}

\vspace{0.5cm}

\textbf{\# Global Plan}\\
\{global\_plan\}

\vspace{0.5cm}

\textbf{\# Previous Observation}\\
\{observation\} [optional]

\end{tcolorbox}

\begin{tcolorbox}[breakable, title=\textbf{Prompt: Execution Generation - BabyAI}, colback=gray!7, colframe=gray!50!black, boxrule=1pt]
You are an exploration master that wants to finish every goal you are given. 
You are placed in a room and you need to accomplish the given goal with actions. 
For each of your turn, you will be given the observation of the last turn. You should choose from two actions: ``Thought'' or ``Action''. If you choose ``Thought'', you should first think about the current condition and plan for your future actions, and then output your action in this turn. Your output must strictly follow this format: ``Thought: your thoughts. Action: your next action''; If you choose ``Action'', you should directly output the action in this turn. Your output must strictly follow this format: ``Action: your next action''. 

\vspace{0.5cm}

The action list you can take:
\begin{enumerate}
    \item turn right
    \item turn left
    \item move forward
    \item go to $<$obj$>$ $<$id$>$
    \item pick up $<$obj$>$ $<$id$>$
    \item go through $<$door$>$ $<$id$>$: $<$door$>$ must be an open door. 
    \item toggle and go through $<$door$>$ $<$id$>$: $<$door$>$ can be a closed door or a locked door. If you want to open a locked door, you need to carry a key that is of the same color as the locked door.
    \item toggle: there is a closed or locked door right in front of you and you can toggle it.
\end{enumerate}
where $<$obj$>$ and $<$id$>$ correspond to objects and index number. 

\vspace{0.5cm}

Reminder:
\begin{enumerate}
    \item The action is restricted to those listed as available. Actions not included in the provided list are considered invalid.
    \item Think when necessary, but prioritize direct action wherever possible throughout the process.
\end{enumerate}

\vspace{0.5cm}

\textbf{\# Example}\\
\{example\}

\vspace{0.5cm}

\textbf{\# Task}\\
\{task\}

\vspace{0.5cm}

\textbf{\# Global Plan}\\
\{global\_plan\}

\vspace{0.5cm}

\textbf{\# Previous Observation}\\
\{observation\} [optional]

\end{tcolorbox}

\begin{tcolorbox}[breakable, title=\textbf{Prompt: Execution Generation -  MAZE}, colback=gray!7, colframe=gray!50!black, boxrule=1pt]
You are an expert maze solver. 
Your objective is to reach the goal in as few steps as possible. 
At each step you will be given information about where the goal is, your current position, and the walls that surround you. 
You should choose from two actions: ``Thought'' or ``Action''. If you choose ``Thought'', you should first think about the current condition and plan for your future actions, and then output your action in this turn. Your output must strictly follow this format: ``Thought: your thoughts. Action: your next action''; If you choose ``Action'', you should directly output the action in this turn. Your output must strictly follow this format: ``Action: your next action''. 
Specifically, 
when you move right, you increase your \texttt{y} position by 1. When you move down, you increase your \texttt{x} position by 1.

\vspace{0.5cm}

The action list you can take:

\begin{enumerate}
    \item move up
    \item move down
    \item move left
    \item move right
\end{enumerate}

\vspace{0.5cm}

Reminder:
\begin{enumerate}
    \item The action is restricted to those listed as available. Actions not included in the provided list are considered invalid.
    \item Think when necessary, but prioritize direct action wherever possible throughout the process.
\end{enumerate}

\vspace{0.5cm}

\textbf{\# Example}\\
\{example\}

\vspace{0.5cm}

\textbf{\# Task}\\
\{task\}

\vspace{0.5cm}

\textbf{\# Global Plan}\\
\{global\_plan\}

\vspace{0.5cm}

\textbf{\# Previous Observation}\\
\{observation\} [optional]

\end{tcolorbox}

\begin{tcolorbox}[breakable, title=\textbf{Prompt: Adherence Degree Judgment (for \Cref{eq:stage1_adherence})}, colback=gray!7, colframe=gray!50!black, boxrule=1pt]
You are an expert in agent tasks. 
You are tasked with evaluating the agent’s execution of a given global plan. Specifically, you are to assess the degree of compliance between the agent’s actions and the strategic guidance outlined in the global plan. Rate it from 0 to 2 points, and explain the reason. \\

\textbf{\textit{\underline{2 Point Answer Criteria}}}: \\
The agent’s execution strictly adheres to the guidance provided in the global plan. All actions are logically aligned with the plan’s objectives and are carried out as instructed. \\

\textbf{\textit{\underline{1 Point Answer Criteria}}}: \\
The agent’s execution demonstrates a partial alignment with the global plan, allowing for minor deviations. For example, in cases where the plan suggests the use of multiple tools, the agent may use at least one relevant tool to support the execution, as long as it does not contradict the overall guidance. \\

\textbf{\textit{\underline{0 Point Answer Criteria}}}: \\
The agent’s execution departs or contradicts the global plan, or contains garbled characters, format errors, disorder, and irrelevant information.

\vspace{0.5cm}

\textbf{\# Task}\\
\{task\}

\vspace{0.5cm}

\textbf{\# Global Plan}\\
\{global\_plan\}

\vspace{0.5cm}

\textbf{\# Execution Step Index}\\
\{execution\_step\_index\}

\vspace{0.5cm}

\textbf{\# Agent Action}\\
\{agent\_action\}

\hdashrule{\textwidth}{1pt}{3pt}

\vspace{0.5cm}

\textbf{Output Format:}  
\begin{verbatim}
```json
{
    "score": xxx,
    "reason": "..."
}
'''
\end{verbatim}

\end{tcolorbox}

\begin{tcolorbox}[breakable, title=\textbf{Prompt: E2E Performance Judgment (for \Cref{eq:stage1_e2e})}, colback=gray!7, colframe=gray!50!black, boxrule=1pt]
You are an expert in agent tasks. 
Please evaluate the end-to-end (E2E) performance of the agent during its interaction with a given environment. The goal is to assess whether the agent accomplishes the target task efficiently and directly, without unnecessary detours or redundancies. 
Rate it from 0 to 2 points, and explain the reason. \\

\textbf{\textit{\underline{2 Point Answer Criteria}}}: 
\begin{enumerate}
    \item The agent successfully completes the task in a direct and efficient manner.
    \item There are no unnecessary steps or redundant actions in the interaction trajectory.
\end{enumerate}

\textbf{\textit{\underline{1 Point Answer Criteria}}}: 
\begin{enumerate}
    \item The task is ultimately completed, but the process includes some level of redundancy or unintended topic drift. 
    \item While the final objective is met, there may be deviations from the optimal path.
\end{enumerate}

\textbf{\textit{\underline{0 Point Answer Criteria}}}: 
\begin{enumerate}
    \item The agent fails to achieve the final task objective.
    \item Contains significant deviations, errors, or inability to progress towards the goal.
\end{enumerate}

\vspace{0.5cm}

\textbf{\# Task}\\
\{task\}

\vspace{0.5cm}

\textbf{\# Agent-Environment Interaction}\\
\{accumulated\_context\}

\vspace{0.5cm}

\textbf{\# Reference Interaction}\\
\{ref\_interaction\}

\hdashrule{\textwidth}{1pt}{3pt}

\vspace{0.5cm}

\textbf{Output Format:}  
\begin{verbatim}
```json
{
    "score": xxx,
    "reason": "..."
}
'''
\end{verbatim}

\end{tcolorbox}

\begin{tcolorbox}[breakable, title=\hypertarget{E2E}{\textbf{Prompt: E2E Performance Evaluation (\textit{LLM-as-Judge})}}, colback=gray!7, colframe=gray!50!black, boxrule=1pt]
You are an expert in agent tasks. 
Please evaluate the end-to-end (E2E) performance of the agent during its interaction with a given environment, focusing on two key dimensions:
\begin{itemize}[leftmargin=*]
    \item \textbf{Task Success}: Did the agent achieve the final goal?
    \item \textbf{Interaction Efficiency}: Was the path direct, logical, and free of redundancy or detours?
\end{itemize}
Assign a score from 0 to 100 and provide a clear justification. Please use the following criteria, and explain the reason. 

\vspace{0.5em}
\textbf{\textit{\underline{90–100: Highly Successful and Efficient.}}} \\ 
The agent demonstrates near-optimal behavior. All of the following must hold: 
\begin{enumerate}
    \item The final task objective is fully and correctly completed.
    \item The interaction trajectory is direct and logically structured.
    \item There are no redundant, repetitive, or off-topic actions.
    \item Any minor errors (e.g., phrasing) do not impede progress.
\end{enumerate}

\textbf{\textit{\underline{70–89: Successful but Inefficient.}}} \\ 
The task is completed, but with non-critical inefficiencies. At least one of the following applies: 
\begin{enumerate}
    \item The agent takes unnecessary steps or detours before completing the task.
    \item There are minor errors or invalid actions that require recovery.
    \item Brief topic drift or redundant reasoning occurs but is self-corrected.
    \item The overall strategy works but is suboptimal in efficiency.
\end{enumerate}

\textbf{\textit{\underline{50–69: Partially Successful.}}} \\ 
Significant progress is made, but the task is not fully completed or requires excessive effort. At least one of the following applies: 
\begin{enumerate}
    \item The agent fails to reach the final goal, but completes most subtasks.
    \item Completion requires major detours, repeated failures, or external hints.
    \item The agent shows inconsistent planning or weak strategy.
    \item Some actions are irrelevant or counterproductive, though not fatal.
\end{enumerate}

\textbf{\textit{\underline{0–49: Failure or Severely Flawed.}}} \\ 
The agent fails to make meaningful progress. At least one of the following applies: 
\begin{enumerate}
    \item The agent fails to complete the core objective.
    \item There is complete task misunderstanding or topic deviation.
    \item The interaction contains repeated invalid actions or infinite loops.
    \item The agent shows no coherent strategy and cannot recover from errors.
\end{enumerate}

\vspace{0.5cm}

\textbf{\# Task}\\
\{task\}

\vspace{0.5cm}

\textbf{\# Agent-Environment Interaction}\\
\{accumulated\_context\}

\vspace{0.5cm}

\textbf{\# Reference Interaction}\\
\{ref\_interaction\}

\hdashrule{\textwidth}{1pt}{3pt}

\vspace{0.5cm}

\textbf{Output Format:}  
\begin{verbatim}
```json
{
    "score": xxx,
    "reason": "..."
}
'''
\end{verbatim}

\end{tcolorbox}


\section{Case Studies}

\begin{figure*}[t]
    \centering
    \includegraphics[width=1\linewidth]{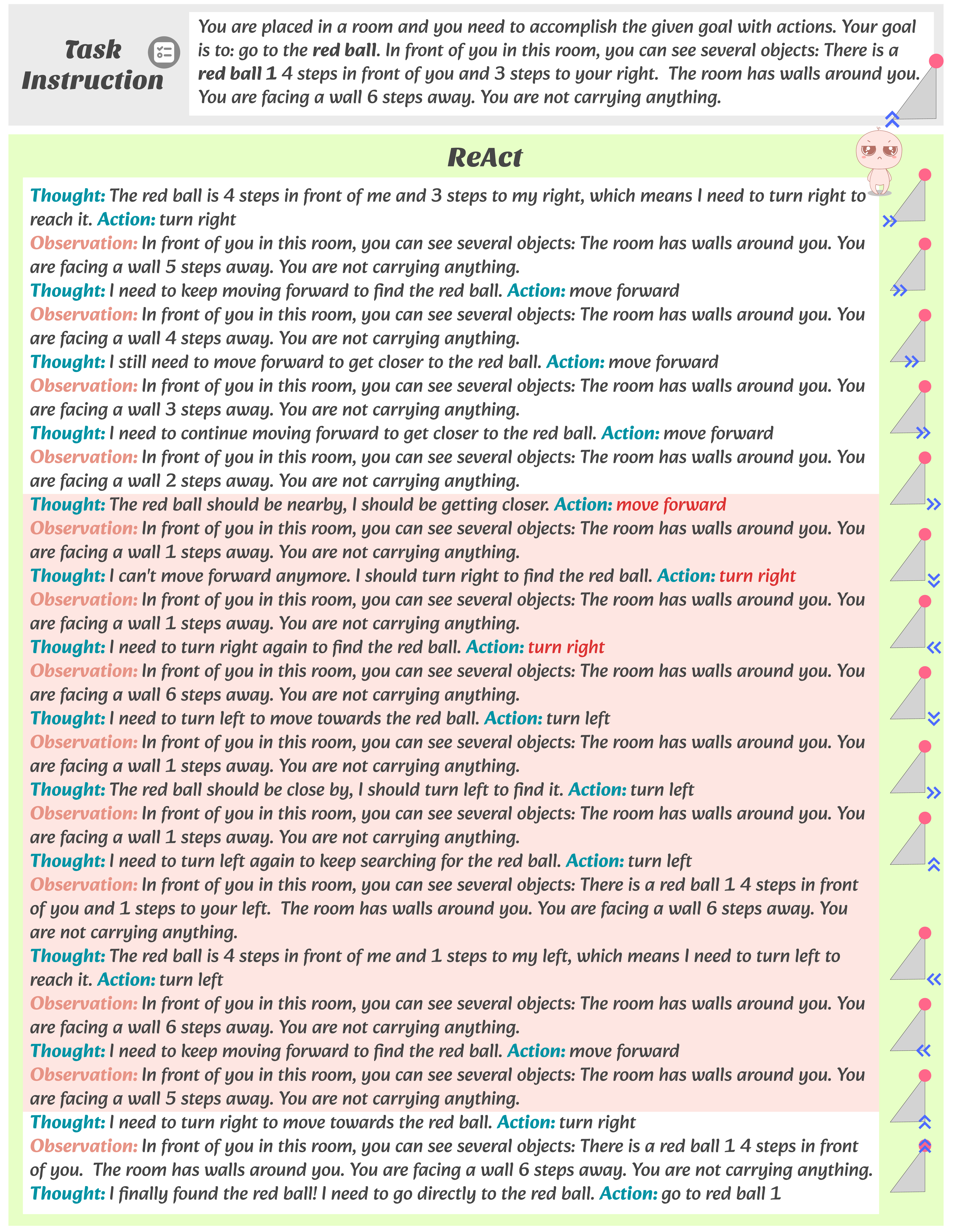}
    \caption{Case study of ReAct~\citep{yao2023react} on BabyAI~\citep{chevalier-boisvert2018babyai}.}
    \label{fig:case_study_ReAct}
\end{figure*}

\begin{figure*}[t]
    \centering
    \includegraphics[width=1\linewidth]{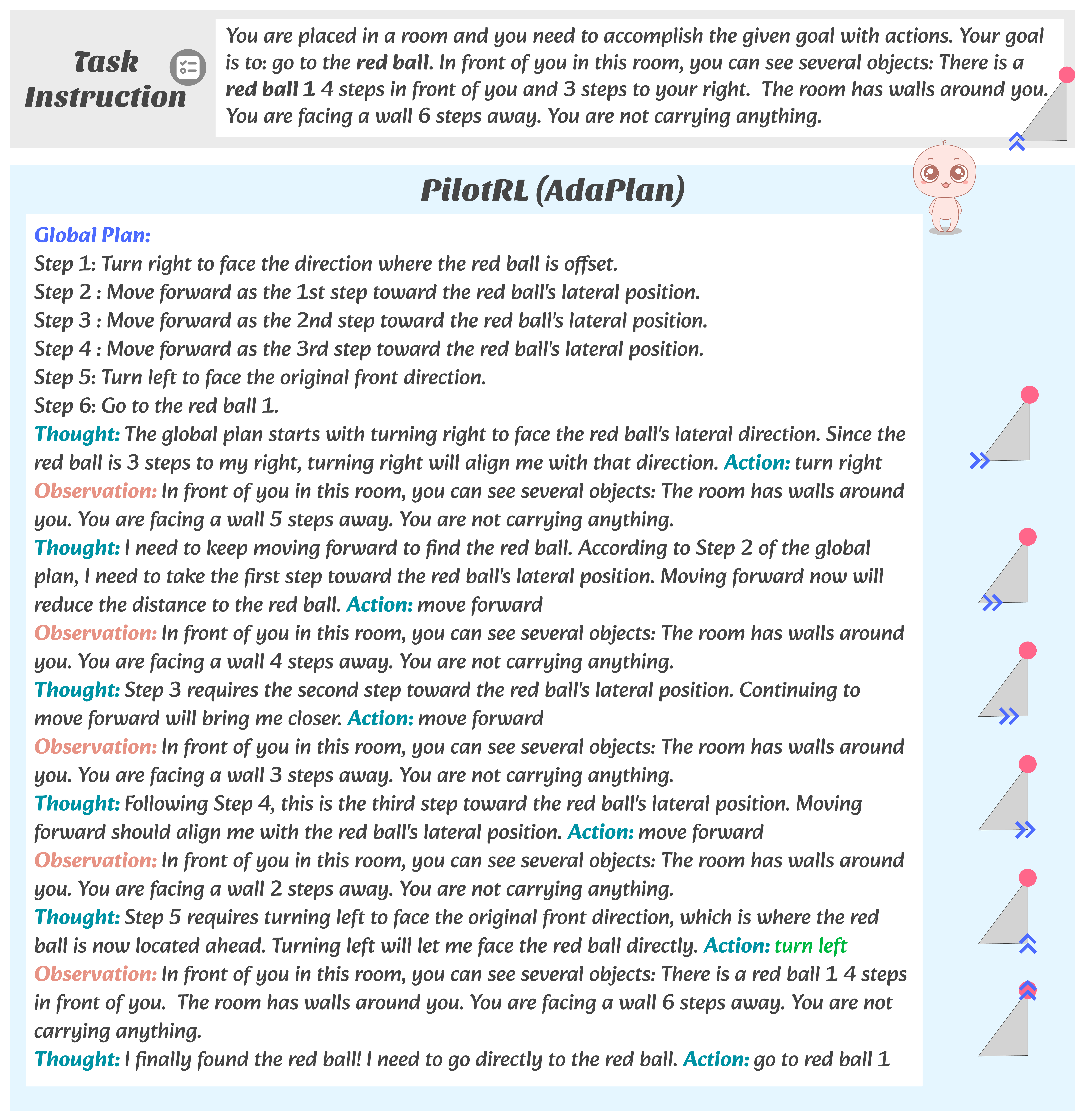}
    \caption{Case study of \ours (AdaPlan) on BabyAI~\citep{chevalier-boisvert2018babyai}.}
    \label{fig:case_study_PilotRL}
\end{figure*}

For agent tasks involving multi-step decision-making, 
generating a global plan to guide the execution of each step is crucial. 
This is because models may forget the previous context after executing multiple steps, leading to redundant actions or failure to accomplish the task. 
As shown in \Cref{fig:case_study_ReAct}, the red annotations indicate redundant interaction trajectories during the problem-solving process. When the agent has already moved three steps to the right, it forgets that the red ball should be directly on its left and continues to move forward, resulting in a large amount of redundant executions. 
In contrast, as depicted in \Cref{fig:case_study_PilotRL}, with the guidance of a global plan, the agent can clearly recognize its relative position of the target, thereby efficiently completing the task.



\end{document}